\documentclass[runningheads]{llncs}
\usepackage{graphicx}
\usepackage{amsmath,amssymb} 
\usepackage{color}

\usepackage[pagebackref=false,breaklinks=true,letterpaper=true,colorlinks,bookmarks=false]{hyperref}

\usepackage{stmaryrd,bm,abraces}
\usepackage{color}
\usepackage{float}
\usepackage{array}
\usepackage{makecell}
\usepackage{multirow}
\usepackage{verbatim}
\usepackage{soul}
\usepackage{lipsum}
\usepackage{float}
\usepackage{enumitem}
\usepackage{arydshln}
\usepackage{mathtools}
\usepackage{epstopdf}
\usepackage{svg}
\usepackage{tikz}
\usepackage{cite}
\usepackage{notoccite} 
\usepackage[lined,ruled,commentsnumbered]{algorithm2e}

\newcolumntype{C}[1]{>{\centering\arraybackslash}p{#1}}

\newcolumntype{K}{>{\centering\arraybackslash}m{2cm}}
\newcolumntype{L}{>{\centering\arraybackslash}m{3cm}}
\newcolumntype{N}{>{\centering\arraybackslash}m{0.01cm}}
\newcolumntype{w}{>{\centering\arraybackslash}m{1.2cm}}
\newcolumntype{Z}{>{\centering\arraybackslash}m{0.1cm}}
\newcolumntype{p}{>{\centering\arraybackslash}m{1.2cm}}
\newcolumntype{v}{>{\centering\arraybackslash}m{1.0cm}}
\newcolumntype{i}{>{\centering\arraybackslash}m{1.8cm}}
\newcolumntype{X}{>{\centering\arraybackslash}m{3.2cm}}
\newcolumntype{Y}{>{\centering\arraybackslash}m{2.7cm}}
\newcolumntype{S}{>{\centering\arraybackslash}m{3.0cm}}
\newcolumntype{F}{>{\centering\arraybackslash}m{4.0cm}}

\newcommand{\etal}{\textit{et al}.}
\newcommand{\etc}{\textit{etc}.}
\newcommand{\ie}{\textit{i}.\textit{e}.}
\newcommand{\eg}{\textit{e}.\textit{g}.}
\newcommand{\vs}{\textit{vs. }}

\newcommand{\fc}[1] {\textcolor{black}{#1}}

\newcommand{\ecochash}{$\mathsf{HBMP}$}
\newcommand{\gist}{$\mathsf{GIST}$}
\newcommand{\vggf}{$\mathsf{VGG}$-$\mathsf{F}$}
\newcommand{\alex}{$\mathsf{AlexNet}$}
\newcommand{\cifar}{$\mathsf{CIFAR}$-$\mathsf{10}$}
\newcommand{\nus}{$\mathsf{NUSWIDE}$}
\newcommand{\imnet}{$\mathsf{ImageNet100}$}
\newcommand{\labelme}{$\mathsf{LabelMe}$}
\newcommand*{\blueline}{\textcolor{blue}{\rule[0.5ex]{1em}{2pt}}}
\newcommand*{\redline}{\textcolor{red}{\rule[0.5ex]{1em}{2pt}}}
\newcommand{\redlined}{\raisebox{2pt}{\tikz{\draw[-,red,dotted,line width = 1.5pt](0,0) -- (4.4mm,0);}}}
\DeclareMathOperator*{\argmin}{arg\,min}
\DeclareMathOperator*{\argmax}{arg\,max}
\usepackage{environ}
\NewEnviron{lequation}{%
\begin{equation}
\scalebox{1.1}{$\BODY$}
\end{equation}
}

\begin{document}
\title{Hashing with Binary Matrix Pursuit} 

\titlerunning{Hashing with Binary Matrix Pursuit}
%
\author{Fatih Cakir\inst{1} \and
Kun He\inst{2} \and
Stan Sclaroff\inst{2}}
%
\authorrunning{Fatih Cakir, Kun He, and Stan Sclaroff}
%

\institute{FirstFuel Software, Lexington, MA \\ \email{fcakirs@gmail.com} \and Department of Computer Science\\
Boston University, Boston, MA \\
\email{\{hekun,sclaroff\}@cs.bu.edu}}
\maketitle              
\begin{abstract}
We propose theoretical and empirical improvements for two-stage hashing methods.
We first provide a theoretical analysis on the quality of the binary codes and show that, under mild assumptions, a residual learning scheme can construct binary codes that fit \fc{any} neighborhood structure with arbitrary accuracy. 
Secondly, we show that with high-capacity hash functions such as CNNs, binary code inference can be greatly simplified for many standard neighborhood definitions, yielding smaller optimization problems and more robust codes.
Incorporating our findings, we propose a novel two-stage hashing method that significantly outperforms previous hashing studies on widely used image retrieval benchmarks.

\end{abstract}

\section{Introduction}


A main challenge for ``learning to hash" methods lies in the discrete nature of the problem. 
Most approaches are formulated as non-linear mixed integer programming problems which are computationally intractable. 
Common optimization remedies 
include discarding the binary constraints and solving for continuous embeddings \cite{SSH, Lai_CVPR2015, cakir2015AdaptHash, hashnet, He_2018_CVPR}. At test time the embeddings are typically thresholded to obtain the desired binary codes. However, even the relaxed problem is highly non-convex requiring nontrivial optimization procedures (\eg, \cite{SHK}), and the thresholded embeddings are prone to large quantization errors, necessitating additional measures (\eg, \cite{Liu_2016_CVPR}).

One prominent alternative to the relaxation approach is \emph{two-stage hashing}, which decomposes the optimization problem into two stages:  binary code inference (i) and  hash function learning (ii). 
For a training set, binary codes are inferred in the inference stage, which are then used as target vectors in the hash function learning stage.
Such methods closely abide to the discrete nature of the problem as the binary codes are directly incorporated into the optimization procedure. 
In two-stage hashing, most of the attention is drawn to the more challenging binary code inference step. 
Typically, this task is itself decomposed into a stage-wise problem where binary codes are learned in an iterative fashion. 
While theoretical guarantees 
for the underlying iterative scheme are usually provided, the overall quality of the binary codes is often overlooked. 
\fc{It is desirable to also determine the quality of the constructed binary codes.}

In this paper, our first contribution is to provide an analysis on the quality of learned binary codes in two-step hashing.
We focus on the frequently considered matrix fitting formulation (\eg, \cite{SHK, Xia_AAAI2014, FastHash, SHK_eccv, sdf_hashing}), 
in which a ``neighborhood structure'' is defined through an affinity matrix and the task is to generate binary codes so as to preserve the affinity values. 
We first demonstrate that ordinary Hamming distances are unable to fully preserve the neighborhood. 
Then, with a weighted Hamming metric, we prove that a {residual learning} scheme can construct binary codes that can preserve any neighborhood with arbitrary accuracy under mild assumptions. 
Our analysis reveals that distance scaling, as well as fixing the dimensionality of the Hamming space, which are often employed in many hashing studies \cite{BRE, MLH, SHK, Wang_ACCV2016}, are both unnecessary.

\fc{
On the other hand, one common inconvenience in two-stage hashing methods is that, 
steps (i) and (ii) are often interleaved,
so as to enable bit correction during training \cite{GraphCut, Shen_CVPR2015, sdf_hashing}.
Bit correction has shown to improve retrieval performance, especially when 
the hash mapping constitutes simple functions such as linear hyperplanes and decision stumps 
\cite{FastHash}.
In contrast, we show that such an interleaved process is unnecessary with high capacity hash functions such as Convolutional Neural Networks (CNNs). 
}

A further benefit of removing interleaving is that the affinity matrix can be constructed directly according to the definition of the neighborhood structure, instead of the pairwise similarities between training instances.
For example, when preserving semantic similarity, the neighborhood is generally defined through class label agreement. Defining the affinity with respect to labels rather than instances yields a much smaller optimization problem for the inference task (i), and provides robustness for the subsequent hash function learning (ii).
In contrast, instance-based inference schemes result in larger optimization problems, often necessitating subsampling to reduce the scale.

With these insights in mind, we implement our novel two-stage hashing method with  standard CNN architectures, and conduct experiments on multiple image retrieval datasets. 
The affinity matrix in our formulation may or may not be derived from class labels, and can constitute binary or multi-level affinities. 
In fact, we consider a variety of experiments that include multi-class ({\cifar} \cite{krizhevsky2009learning}, {\imnet} \cite{nus-wide-civr09}), multi-label ({\nus} \cite{deng2009imagenet}) and unlabeled ($\mathsf{22K}$ {\labelme} \cite{russell2008labelme}) datasets.
We achieve new state-of-the-art performance for all of these datasets.
In summary, our contributions are: 
\begin{enumerate}
\item We provide a technical analysis on the quality of the inferred binary codes demonstrating that under mild assumptions
we can fit any neighborhood with arbitrary accuracy. Our analysis is relevant to the formulations used in many two-stage hashing methods (\eg, \cite{TSH, FastHash, Xia_AAAI2014, sdf_hashing, Zhuang_2016_CVPR}).
\item We demonstrate that with high-capacity hash functions such as CNNs, the bit correction task is expendable. 
As a result, binary code inference can be performed on items that directly define the neighborhood, yielding more robust target vectors and improving the retrieval performance.
We achieve state-of-the-art performance in four standard image retrieval benchmarks.
\end{enumerate}



\section{Related Work}
\label{related-work}
We only review hashing studies most relevant to our problem. For a general survey, please refer to \cite{hash_survey2018}.

The two-stage strategy for hashing was pioneered by Lin \etal~\cite{TSH} in which the authors reduced the binary code inference task
into a series of binary quadratic programming (BQP) problems. The target codes are optimized in an iterative fashion and traditional machine learning classifiers such as Support Vector Machines (SVMs) and linear hyperplanes that fit the target vectors are employed as the hash functions. 
In \cite{FastHash}, the authors proposed a graph-cut algorithm to solve the BQP problem and employed boosted decision trees as the hash functions. The graph-cut algorithm has shown to yield a solution well bounded with respect to the optimal value \cite{Boykov_graphcut_2001}. The authors also demonstrated that, with shallow models an interleaved process of binary code inference and hash function learning allowed bit correction and improved the retrieval performance. Differently, Xia \etal~\cite{Xia_AAAI2014} proposed using a coordinate descent algorithm with Newton's method to solve the BQP problem and utilized CNNs as the hash mapping. Do \etal~\cite{sdf_hashing} solved the the BQP problem using semidefinite relaxation and Lagrangian approaches. They also investigate the quality of the relaxed solution and prove that it is within a factor of the global minimum. Zhuang \etal~\cite{Zhuang_2016_CVPR} demonstrated that the same BQP approach can be extended to solve a triplet-based loss function. 
Other work reminiscent of these two-stage methods include hashing techniques that employ alternating optimization to minimize the original optimization problem \cite{DGH_nips2014, GraphCut, Shen_CVPR2015, SHK_eccv}. 

While error-bounds and convergences properties of the underlying iterative scheme is usually provided, none of the aforementioned studies provide a technical guarantee on the overall quality of the constructed binary codes. In this study we provide such an analysis. 
Our technical analysis has connections to low-rank matrix learning \cite{Shalev_Shwartz_ICML2011, Zhang_NIPS2012, matrix_completion, greedy_rank_one} in which we construct binary codes in a gradient descent or \textit{matrix pursuit} methodology.
Differently, we constrain ourselves with binary rank-one matrices, which are required for Hamming distance computations. Also, while not all two-stage hashing studies follow an interleaved process (\eg, \cite{STH, TSH, aaai_hash}), to the best of our knowledge, all construct the affinity matrix using training instances. This warrants an in-depth look to the necessity of such a process when high-capacity hash functions are employed.

Our hashing formulation follows the matrix fitting formulation which is almost exclusively used in two-stage methods. This formulation was originally proposed in \cite{SHK} and has been widely adopted in subsequent hashing studies (\eg, \cite{TSH, FastHash, SmartHashing, Xia_AAAI2014, sdf_hashing}). Whereas the major contribution in this paper lies in establishing convergence properties of the binary code inference task, our formulation also has subtle and key differences to \cite{SHK} and other two-stage methods. Specifically, we allow weighted hamming distances with optimally learned weights given the inferred binary codes. We perform inference directly on items that define the neighborhood, enabling more robust target vector construction as will be shown.  
In retrieval experiments, we compare against recent hashing studies, including \cite{SH, ITQ, Shen_CVPR2015, Wang_SPLH_ICML2010, Ziming_VeryDeep_CVPR2016, lin2016structured, Wang_ACCV2016, Li_IJCAI2016, mihash, hashnet}, and achieve state-of-the-art performances. 

\setlength{\belowdisplayskip}{5pt} \setlength{\belowdisplayshortskip}{2pt}
\setlength{\abovedisplayskip}{5pt} \setlength{\abovedisplayshortskip}{2pt}
\section{Formulation}
In this section, we first discuss the two stages of our hashing formulation: binary code inference and hash mapping learning. An analysis on affinity matrix construction comes next. All proofs are provided in the supplementary material.
\subsection{Binary Code Inference}
\label{sec:binary}
In this section, we explain our inference step (i). We are given a metric space $(\mathcal{X}, d)$ where $\mathcal{X}=\{\mathbf{x}_1,\cdots, \mathbf{x}_n\}$ denotes a set of items and $d:\mathcal{X} \times \mathcal{X} \to \mathbb{R}_{\geq 0}$ is a metric. 
Note that $\mathbf{x}$ can correspond to instances, labels, multi-labels or any item that is involved in defining the neighborhood. 
Given the assumption that the neighborhood is defined through metric $d$, we 
learn the hash mapping $\Phi:\mathcal{X}\to\mathbb{H}^b$ by optimizing the 
\textit{neighborhood preservation fit}:
\begin{equation}
\min_{\Phi}\sum_{i,j} [\gamma d(\mathbf{x}_i, \mathbf{x}_j) -d_h(\Phi(\mathbf{x}_i), \Phi(\mathbf{x}_j))]^2,
\label{main_obj}
\end{equation}
where $d_h$ is the Hamming distance and $\gamma$ is a suitably selected scaling parameter. In order to scale distances to the range of $d_h$, we set $\gamma={b}/{d_{\max}}$ where $d_{\max} = \max_{\mathbf{x}, \mathbf{y} \in \mathcal{X}} d(\mathbf{x}, \mathbf{y})$ is known.\footnote{Our later analysis in this section only requires $d_{\max}$ to be bounded.}
Solving Eq.~\ref{main_obj} entails discrete loss minimization,
which in general is a non-linear mixed-integer programming problem. Instead, two-stage methods decompose the solution 
into two steps, the first involving a binary integer program to find a set of binary codes, or auxilliary variables $\{\mathbf{u}_i \in \mathbb{H}^b\}_{i=1}^n$ that minimize Eq.~\ref{main_obj}. 
This program can be formulated as:
\begin{equation}
\min_{\mathbf{u}} \sum_{i,j } [ \gamma d(\mathbf{x}_i, \mathbf{x}_j) - d_h(\mathbf{u}_i, \mathbf{u}_j)]^2 
= \min_{\mathbf{u}}\frac{1}{4} \sum_{i,j } [\mathbf{u}_i^\top \mathbf{u}_j - s(\mathbf{x}_i, \mathbf{x}_i)]^2,
\label{main_obj2}
\end{equation} 
where $ s(\mathbf{x}_i, \mathbf{x}_j) = b- 2 \gamma d(\mathbf{x}_i, \mathbf{x}_j), \forall i,j \in \mathcal{X}$. 
While the LHS of Eq.~\ref{main_obj2} is a distance equivalence problem, the RHS is an affinity matching task. Such affinity based preservation objectives have also been considered previously
\cite{SH, SSH, SHK, Wang_ACCV2016}.


In our formulation, we consider weighted Hamming distances by weighting each bit in $\mathbf{u}$. 
The weighted Hamming distance has been used in past studies to provide more granular similarities compared to its unweighted counterpart 
(\eg, \cite{WHD_AnnoSearch_CVPR_2006, MDH, CGHASH, WHD_CVPR_2013, WHD_MM_2013}).
While this hashing scheme still enjoys low memory footprint and fast distance computations, weighting the individual bits enables us to construct binary codes that better preserve affinity values, as will be shown later.

We reformulate Eq.~\ref{main_obj2} by defining weight vector
$\bm{\alpha} = [\mathbf{\alpha}_1, \cdots, \mathbf{\alpha}_b]^\top$:
\begin{equation}
\frac{1}{4} \sum_{i,j } [(\bm{\alpha} \odot\mathbf{u}_i)^\top \mathbf{u}_j - s(\mathbf{x}_i, \mathbf{x}_i)]^2 \propto \frac{1}{2} \|\mathcal{U} - \mathcal{R}\|_F^2 = f(\mathcal{U}),
\label{main_obj3}
\end{equation}
where $\odot$ denotes the Hadamard product, $\mathcal{U}_{ij} = (\bm{\alpha} \odot\mathbf{u}_i)^\top \mathbf{u}_j , \mathcal{R}_{ij} =  s(\mathbf{x}_i, \mathbf{x}_j), \forall i,j \in \mathcal{X}$ and $\|\cdot\|_F$ denotes the Frobenius norm.
We note that the \textit{affinity matrix} $\mathcal{R}$ is real and symmetric as per its construction from metric $d$. 

Let $\mathbf{V} = [\mathbf{u}_1, \cdots, \mathbf{u}_n]^\top \in \mathbb{H}^{n \times b}$ denote the \textit{binary code matrix}, then $\mathcal{U}$ can be written as the weighted sum of $b$ rank-one matrices $\sum_{k=1}^b \alpha_k \mathbf{v}_k\mathbf{v}_k^\top$ where $\mathbf{v}_k \in \{-1, 1\}^n$ is the $k$-th column in $\mathbf{V}$. Given this fact, our binary inference problem can be reformulated as:
\begin{equation}
\begin{split}
& \min ~f(\mathcal{U}), 
 ~~\text{s.t.}~~\mathcal{U}  = \sum_{k=1}^b \alpha_k \mathbf{v}_k\mathbf{v}_k^\top , ~ \mathbf{v} \in \{-1,+1\}^n.
\end{split}
\label{main_obj4}
\end{equation}
The additive property of $\mathcal{U}$ is attractive, since it suggests that the problem could be solved by a \emph{stepwise} algorithm that adds the $\mathbf{v}_k$'s one by one.
In particular, we will apply the projected gradient descent algorithm to solve Eq.~\ref{main_obj4}. 
Starting with an initial value, $\mathcal{U}_0 = \mathbf{0}$, an update step can be formulated as:
\begin{equation}
\mathcal{U}_t \leftarrow \mathcal{U}_{t-1} + \alpha_t \mathbf{v}_t\mathbf{v}_t^\top,
\label{sgd}
\end{equation}
where
\begin{equation}
\mathbf{v}_t = \argmax_{\mathbf{v} \in \{-1,+1\}^n} \langle\mathbf{v}\mathbf{v}^\top, -\nabla f(\mathcal{U}_{t-1})\rangle 
\label{gradient}
\end{equation}
finds the projection of the negative gradient direction $-\nabla f(\mathcal{U}_{t-1})$ in the subspace spanned by rank-one binary matrices, and $\alpha_t$ is a step size. 
This projection is important for maintaining the additive property in Eq.~\ref{main_obj4}.

Since $\langle\mathbf{v}\mathbf{v}^\top, \nabla f \rangle = \mathbf{v}^\top \nabla f\mathbf{v}$, Eq.~\ref{gradient} is a BQP problem which in general is {NP-hard}. 
Here, we take a spectral relaxation approach which is also used in past methods (\eg, \cite{SH, SHK, TSH}). 
A closed-form solution to Eq.~\ref{gradient} exists if the binary vector $\mathbf{v}$ is relaxed to continuous values. Specifically, if $Q = -\nabla f(\mathcal{U})$, 
the following relaxation yields the Rayleigh Quotient \cite{Horn_Matrix_Analysis_1983}:
\begin{equation}
\max_{\mathbf{v}^\top \mathbf{v} = n} \mathbf{v}^\top Q\mathbf{v} = n\lambda_{\max}(Q),
\label{rayleigh}
\end{equation}
where $\lambda_{\max}$ denotes the largest eigenvalue, and the optimal solution, $\mathbf{v}^*$, is the corresponding eigenvector. The binarized value of $\mathbf{v}^*$, $\text{sgn}(\mathbf{v}^*)$, is an approximate solution for Eq.~\ref{gradient}. This solution can optionally be used as an initial point for BQP solvers in further maximizing Eq.~\ref{gradient}, (\eg, \cite{Merz_BQP_2002, Buchheim_BQP_2013, Wang_BQP_2017}). 
Note that the main technical results to be given are independent of the particular BQP solver. 

The negative gradient $-\nabla f(\mathcal{U}_{t-1}) = \mathcal{R} - \sum_{k=1}^{t-1} \mathbf{v}_k\mathbf{v}_k^\top $, also a symmetric matrix, can be considered as the \textit{residual} at iteration $t-1$. At each iteration, we find the most correlated rank-one matrix with this residual and move our solution in that direction. If the step size $\alpha_t$ is set to $1$ for all $t$, then $\mathcal{U}$ can be decomposed as the product of the binary code matrices $\mathbf{V}\mathbf{V}^\top$, yielding ordinary Hamming distances. However, with constant step sizes, the below property states that there exist certain affinity matrices $\mathcal{R}$ such that no $\mathcal{U}$ exists that \textit{fits} $\mathcal{R}$. 

\medskip
\noindent \textbf{Property 1.} \textit{Let $Q_{t}$ be the residual $-\nabla f(\mathcal{U}_{t})$ at iteration $t$. There exists a $\mathcal{R}$ such that $\forall t, \|Q_{t}\|_F > 0$}.%


\smallskip
Such a result motivates us to relax the constraint on the step size parameter $\alpha_t$. If $\alpha$ is relaxed to any real value, then what we have essentially is weighted Hamming distances and we demonstrate that one can monotonically decrease the residual $\mathcal{R}$ in this case. 
We now provide our main theorem:

\medskip
\noindent \textbf{Theorem 2.} \textit{If $\alpha_t \in \mathbb{R}$, then the gradient descent algorithm Eq.~\ref{sgd} - \ref{gradient} satisfies}
\begin{equation}
\|Q_t\|_F \leq \eta^{t-1}\|Q_{t-1}\|_F,~~\forall t
\label{theorem}
\end{equation}
\textit{where $\eta \in [0, 1]$}. 



\smallskip
Theorem 2 states that the norm of the residual is only monotonically non-increasing. However, it may not strictly decrease, since the solution $\mathbf{v}_t$ of Eq.~\ref{gradient} can actually be orthogonal to the gradient, \ie, $\mathbf{v}_t^\top Q_{t-1}\mathbf{v}_t$ might be zero. 
If we ensure non-orthogonal {directions} are selected at each iteration, then the residual strictly decreases, as the following corollary states.

\medskip
\noindent \textbf{Corollary 3.} \textit{If $\mathbf{v}_t^\top Q_{t-1} \mathbf{v}_t \neq 0, ~\forall t$ then the residual norm $\|Q_t\|_F$ strictly decreases}.

\smallskip
Although the directions $\mathbf{v}_t\mathbf{v}_t^\top$ are greedily selected with step sizes $\alpha_t$, one can refine step sizes of all past directions at each iteration. This generally leads to much faster convergence. More formally, we can refine the step size parameters by solving the following regression problem:
\begin{equation}
\bm{\alpha}^* = \argmin_{{\alpha_1, \cdots, \alpha_t}} \frac{1}{2}\|\sum_{k=1}^t \alpha_k \mathbf{v}_k\mathbf{v}_k^\top - \mathcal{R}\|_F^2.
\label{regression}
\end{equation}
Fortunately, Eq.~\ref{regression} is an ordinary least squares problem admitting a closed-form solution. Let $\mathbf{\widehat{v}}_k = \text{vec}(\mathbf{v}_k \mathbf{v}_k^\top)$ and $\mathbf{\widehat{r}} = \text{vec}(\mathcal{R})$ where vec$(\cdot)$ denotes the vectorization operator. Given $\mathbf{\widehat{V}}_t = [\mathbf{\widehat{v}}_1, \cdots, \mathbf{\widehat{v}}_t]$, the minimizer of Eq.~\ref{regression} is
\begin{equation}
\bm{\alpha}_t^* = (\mathbf{\widehat{V}}_t^\top \mathbf{\widehat{V}}_t)^{-1}\mathbf{\widehat{V}}_t^\top \mathbf{\widehat{r}},
\label{solution-regression}
\end{equation}
where $\bm{\alpha}_t^* = [{\alpha_1^*, \cdots, \alpha_t^*}]^\top$. 
The solution requires $ \mathcal{O}(t^3) + \mathcal{O}(t^2 n^2) +\mathcal{O}(tn^2)$ operations with $n=|\mathcal{X}|$. If $n > \sqrt{t}$, the time complexity is dominated by the $\mathcal{O}(t^2 n^2)$ term. 
Note that in practice, typical values for $t$, the number of bits, are small ($ < 100$) and can be considered a constant factor. 

We now provide a property indicating that this {refinement} of the step-sizes does not break the monotonicity as defined in Theorem 2 and Corollary 3. 

\medskip
\noindent \textbf{Property 4.} \textit{Let $Q_t$ be the residual matrix at iteration $t$ and $\alpha_t$ set according to Theorem 2. Let $\widehat{Q}_t$ be the residual after refining the step-sizes $\bm{\alpha_t} = [\alpha_1, \cdots, \alpha_t]^\top$ using Eq.~\ref{solution-regression}. Then $\|\widehat{Q}_t\|_F \leq \|Q_t\|_F$}.


\smallskip
After learning $\mathcal{U} = \sum_{k=1}^\top \alpha_k \mathbf{v}\mathbf{v}^t = \mathbf{A} \odot \mathbf{V}\mathbf{V}^\top$ where $A_{k,\cdot} = [\alpha_1, \cdots, \alpha_t], \forall k$ we obtain our binary code matrix $\mathbf{V} = [\mathbf{u}_1, \cdots, \mathbf{u}_n]^\top$ that contains the target codes for each element $\{\mathbf{x}_1,\cdots, \mathbf{x}_n\} \in \mathcal{X}$. This ends our inference step (i). We summarize our inference scheme in Alg. {Binary code inference}. 

\smallskip
\noindent \textbf{Remarks.} 
We consider two different binary inference schemes: \texttt{constant} where the binary codes are constructed with constant step sizes yielding ordinary Hamming distances; and, \texttt{regress} where each bit is weighted yielding the weighted Hamming distance. 
For \texttt{regress}, since $(\bm{\alpha} \odot\mathbf{u}_i)^\top \mathbf{u}_j = b(1-2{d(\mathbf{x}_i, \mathbf{x}_j)}/{d_{\max}}) $ in Eq.~\ref{main_obj3}, we can embed the constant $b$ into the weight vector variable $\bm{\alpha}$. As a result, in contrast to hashing methods where the Hamming space dimensionality $b$ must be specified (\eg, to set margin and scaling parameters \cite{SHK, SHK_eccv, Wang_ACCV2016}), our method only requires $d_{\max}$ to be bounded.
On the other hand, regular Hamming distance, or \texttt{constant}, requires scaling with $b$ beforehand. 
The approximate solution of Eq.~\ref{rayleigh} can be improved by using off-the-shelf BQP solvers.
In Alg. Binary code inference, we refer to such solvers as the subroutine \texttt{Improve}$(\cdot)$. In this paper, we consider using a simple heuristic \cite{Merz_BQP_2002}, which merely requires a positive objective value for Eq.~\ref{gradient}.

We now proceed with step (ii): hash mapping learning. 


\setlength{\textfloatsep}{1pt} 
\setlength{\intextsep}{1pt} 
\setlength{\floatsep}{1pt} 
\begin{algorithm}[t!]
\footnotesize
\LinesNumbered\DontPrintSemicolon
\SetKwData{Left}{left}\SetKwData{This}{this}\SetKwData{Up}{up}\SetKwData{Ind}{ind}
\SetKwFunction{Union}{Union}\SetKwFunction{FindCompress}{FindCompress}
\SetKwInOut{Input}{input}\SetKwInOut{Output}{output}
\Input{$\mathcal{X}=\{\mathbf{x}_1, \cdots, \mathbf{x}_n\}$, $d: \mathcal{X} \times \mathcal{X} \to \mathbb{R}_{\geq 0}$. Boolean variable \texttt{regress}. (Optional) Procedure \texttt{Improve($Q, \mathbf{v}_0$)} to improve the solution $ \mathbf{v}^\top Q\mathbf{v}$ s.t. $\mathbf{v} = \{-1,1\}^n$ where $\mathbf{v}_0$ is an initial solution. $\mathcal{U} = \mathbf{0}$.}
\Output{Code matrix $\mathbf{V} = [\mathbf{u}_1, \cdots, \mathbf{u}_n]^\top$, weight vector $\bm{\alpha} = [\alpha_1,\cdots, \alpha_T]^\top$}
\BlankLine
$\gamma = \frac{1}{d_{\max}}$, 
$\mathcal{R}_{ij} =   1- 2 \gamma d(\mathbf{x}_i, \mathbf{x}_j)$
(if \texttt{regress}),
$\mathcal{R}_{ij} =  s(\mathbf{x}_i, \mathbf{x}_j) \times b$ (if $\neg$\texttt{regress}) \;
\For{$t\leftarrow 1,..., T$}{
$\mathbf{v}_t \gets $ eigenvector corresponding to the largest eigenvalue of $-\nabla f(\mathcal{U}_{t-1})$\;
$\mathbf{v}_t \leftarrow \text{sgn} (\mathbf{v}_t)$, $\alpha_t \leftarrow 1$ \;
$\mathbf{v}_t \leftarrow$ \texttt{Improve($-\nabla f(\mathcal{U}_{t-1}), \mathbf{v}_t)$} (optional)\;
\If(\tcp*[h]{$\alpha_t \in \mathbb{R}$}){\texttt{regress}}{
Set $[\alpha_1, \cdots, \alpha_t]^\top$ using Eq.~\ref{solution-regression} \;
}
$\mathcal{U}_t \leftarrow \sum_t \alpha_t \mathbf{v}_t \mathbf{v}_t^\top $, $V_{\cdot, t} \leftarrow \mathbf{v}_t$ 
\tcp{Append $\mathbf{v}_t$ to $\mathbf{V}$}
}
\caption{Binary code inference}
\label{alg:algorithm1}
\end{algorithm}

\subsection{Hash Mapping Learning}
\label{hash-mapping-learning}
Recall that we inferred target codes $\mathbf{u} \in \mathbb{H}^b$ for each item $\mathbf{x} \in \mathcal{X}$, where $\mathbf{x}$ may correspond to data instances, classes, multi-labels \etc, depending on the neighborhood definition.
For example, when the dataset is unsupervised and the neighborhood is defined merely through data instances, then $\mathcal{X}$ may correspond to the feature space with $d(\mathbf{x}_i, \mathbf{x}_j)$ being the Euclidean distance. For multi-class datasets, $\mathcal{X}$ and $d(\mathbf{x}_i, \mathbf{x}_j)$ may represent the set of classes and the distance values between pairs of classes, respectively. For multi-label datasets, $\mathcal{X}$ may correspond to the set of possible label combinations. Our binary inference scheme constructs target codes to items that \textit{directly} define the neighborhood. In the experiments section, we cover various scenarios. 

If $\mathcal{X}$ does not represent the feature space, then after the binary inference step (i), the target codes get assigned to data instances in a one-to-many fashion, depending on the relationship between the target code and data instance. For sake of clarity, we assume $\mathcal{X}$ is the feature space in this section.

We employ a collection of hash functions to learn the mapping, where a function $f:\mathcal{X} \to \{-1, 1\}$ accounts for the generation of a bit in the binary code. Many types of hash functions are considered in the literature. For simplicity, we consider the thresholded scoring function:
\begin{equation}
\label{hash_function}
f(\mathbf{x})\triangleq\text{sgn} (\psi(\mathbf{x})),
\end{equation}
where $\psi$ can be either a shallow model such as a linear function, or a deep neural network. In experiments, we consider both types of embeddings. 
$\Phi (\mathbf{x}) = [f_1 (\mathbf{x}),\cdots,f_b (\mathbf{x})]^\top$ then becomes a vector-valued function to be learned. 

Recall that we inferred target codes $\mathbf{u} \in \mathbb{H}^b$ for each element $\mathbf{x} \in \mathcal{X}$. 
Having the target codes at our disposal, we now would like to find $\Phi$ such that the Hamming distances between $\Phi(\mathbf{x})$ and the corresponding target codes $\mathbf{u}$ are minimized. 
Hence, the objective can be formulated as:
\begin{equation}
\label{main_objective}
\sum_{i=1}^n d_h(\Phi(\mathbf{x}_i),\mathbf{u}_i).
\end{equation}
The Hamming distance is defined as $d_h( \Phi(\mathbf{x}_i), \mathbf{u}_i) = \sum_t [\![f_t(\mathbf{x}_i) \neq u_{it}]\!]$  where both $d_h$ and the functions $f_t$ are non-differentiable. Fortunately, we can relax $f_t$ by dropping the $sgn$ function in Eq.~\ref{hash_function} and derive an upper bound on the Hamming loss. Note that $d_h( \Phi(\mathbf{x}_i), \mathbf{u}_i) = \sum_t [\![f_t(\mathbf{x}_i) \neq u_{it}]\!] \leq \sum_t l(-u_{it} \psi_t(\mathbf{x}_i))$ with a suitably selected convex margin-based function $l$. Thus, by substituting this surrogate function into Eq.~\ref{main_objective}, we can directly minimize this upper bound using stochastic gradient descent.
We use the hinge loss as the upper bound $l$.

As similar to other two-stage hashing methods, at the heart of our formulation are the target vectors which are inferred as to fit the affinity matrix $\mathcal{R}$. Next, we take a closer look on how to construct this affinity matrix. 
\subsection{Affinity Matrix Construction}
The affinity matrix can be defined through pairwise similarities of items that directly define the neighborhood, which may not correspond to training instances.  
Despite this flexibility of the formulation, previous related hashing studies generally consider using training instances.

\begin{figure}[t]
 \centering
     \includegraphics[width=.4\textwidth]{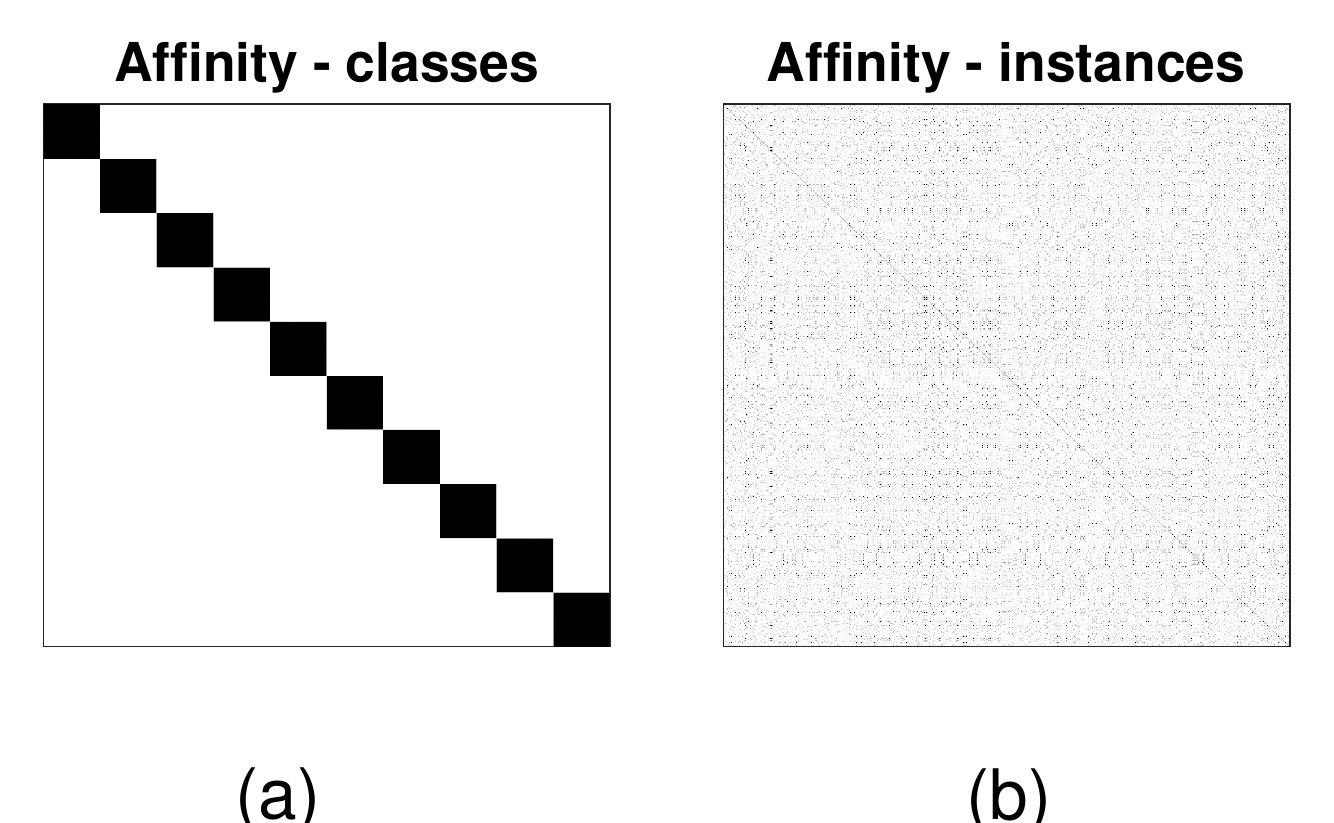}
     
     \includegraphics[width=.8\textwidth]{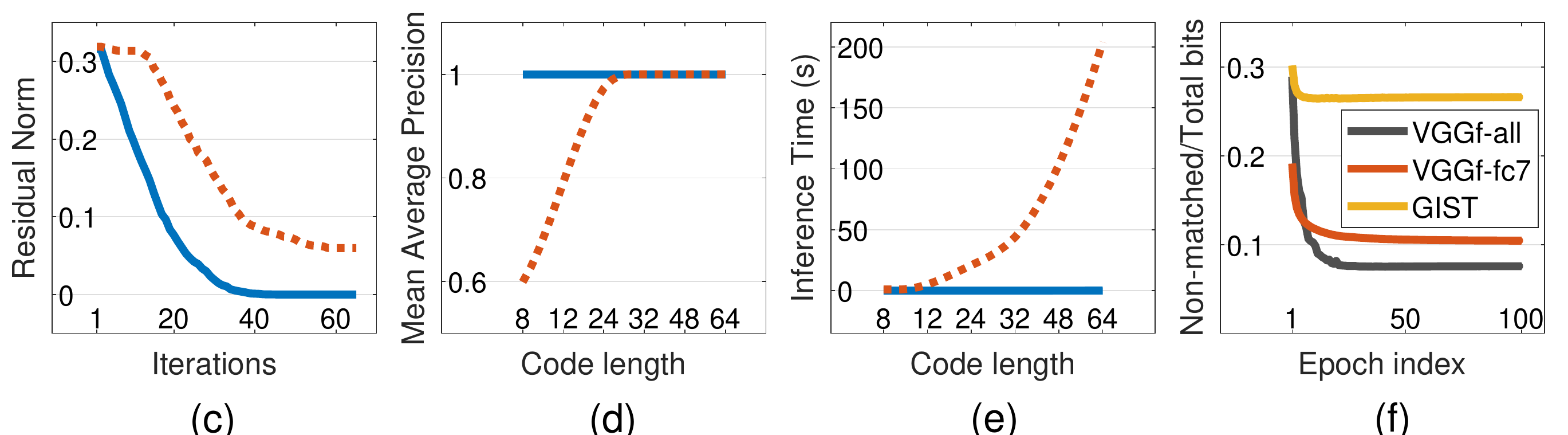}
 \caption{In a series of experiments, we compare two sets of binary codes constructed with two different affinity matrices: class {(a)} and instance based {(b)}. {(c)-(e)} contrasts the binary codes with respect to residual norm, $\mathsf{mAP}$ and inference time. Results for binary codes inferred from the class and instance affinity matrices are denoted with (\blueline) and (\protect \redlined), respectively. We also learn hash functions with varying complexities to fit the inferred binary codes and plot the fraction of non-matched bits to the total number bits {(f)}.}
   \label{fig:aff}
 \end{figure}

For certain neighborhoods, constructing the affinity matrix with training instances might yield suboptimal binary codes. 
To illustrate this case, consider Fig. \ref{fig:aff} where we compare two sets of binary codes inferred from two different affinity matrices in a series of experiments.  
The neighborhood definition in these experiments is a standard one, typically found in nearly all hashing work.
Specifically, we assume 10 classes and define the class affinity matrix as shown in Fig. \ref{fig:aff} (a). 
We also consider a hypothetical set of 1000 instances, each assigned to one of these 10 classes, and construct the affinity matrix as shown in Fig. \ref{fig:aff} (b) which we simply refer as the instance affinity matrix. 
Similarity of the instances are based on their class id's and deduced from the class affinity matrix. 
We infer binary codes under varying lengths as to reconstruct the class and instance affinity matrices. 
As explained in Section \ref{hash-mapping-learning}, instances are assigned the binary code of their respective classes for the class based inference. The experiments are repeated 5 times and average results are reported.

\setlength{\belowcaptionskip}{15pt}

We first highlight the residual matrix $Q_t$ norm in Fig. \ref{fig:aff} (c). 
Note that the residual norm of the class based inference converges to zero with fewer iterations: 40 bit codes are able to reconstruct the class affinity matrix with minimal discrepancy. On the other hand, lengthier codes are required to fully reconstruct the instance affinity matrix. 
We also provide the retrieval performance for the two sets of binary codes. Mean Average Precision ($\mathsf{mAP}$) is the evaluation criterion. For this experiment, 100 instances are sampled from the instance set as queries, while the rest constitute the retrieval set. As demonstrated in Fig. \ref{fig:aff} (d), their is a dramatic difference in $\mathsf{mAP}$ values especially with compact codes. The difference can be as large as 0.40. 
This type of sub-optimality for the binary codes inferred through the instance affinity matrix have also been observed previously (\eg, \cite{Zhuang_2016_CVPR}).
Lastly, Fig. \ref{fig:aff} (e) gives the training time for the two inference schemes. While the inference time depends on the particular BQP solver, the number of decision variables nevertheless scales quadratically with the number of items in $\mathcal{X}$, as seen by the dramatic difference in the training time between the two inference schemes, especially with lengthier codes. Depending on the instance matrix size, the difference can easily scale up requiring subsampling to reduce the scale of the optimization task.

Given the evident disadvantages, why is the affinity matrix constructed from instances? The primary reason is because in most two-stage hashing methods the inference and hash function learning steps are interleaved for \textit{on-the-fly} bit correction purposes. 
This requires the affinity matrix to correspond to pairwise instance similarities as the inferred bits will \textit{immediately} be used for training the hash functions. However, given recent advances in deep learning, high-capacity predictors are becoming available, nullifying the need for bit correction. Consequently, one can opt to solve a smaller and more robust optimization problem defined on items that directly define the neighborhood. 

To illustrate this point we learn hash functions of varying complexities to fit the set of binary codes,\footnote{\scriptsize{These binary codes are obtained from the class based inference scheme, though similar behavior is exhibited with codes obtained with instance based inference.}} and plot the fraction of non-matched bits to total number bits during hash function learning. We use the training set of {\cifar} and train the hash functions to fit the inferred 32-bit binary codes (total: 32$\times$50,000 bits). We consider single layer neural networks on {\gist} \cite{gist} and $\mathsf{fc7}$ features of a {\vggf} network \cite{Chatfield_Return_BMVC14} pretrained on ImageNet \cite{deng2009imagenet}, in addition to fine-tuning all the {\vggf} layers. 
Fig.~\ref{fig:aff} (f) gives the results. Notice that as the capacity of the hash function increases the ratio of non-matched bits decrease significantly. While this ratio is above 0.25 with a single layer neural net on {\gist}, the single layer neural net trained on $\mathsf{fc7}$ features yields just above 10\% unmatched bits. When we fine-tune all layers of a {\vggf} network this percentage reduces well below 10\%. We can induce that with more complex architectures the ratio will diminish even more so. 

We incorporate these insights into our formulation and conduct retrieval experiments against competing methods in the next section, where we achieve new state-of-the-art performances. 

\section{Experiments}
We conduct experiments on widely used image retrieval benchmarks: {\cifar} \cite{krizhevsky2009learning}, 
\nus~\cite{nus-wide-civr09}, $\mathsf{22K}$ \labelme~\cite{russell2008labelme} and \imnet~\cite{deng2009imagenet}. 

\smallskip
\noindent \textbf{\cifar} is a dataset for image classification and retrieval, containing 60K images from 10 different categories. We follow the setup of \cite{Lai_CVPR2015, Zhuang_2016_CVPR, Li_IJCAI2016, Wang_ACCV2016}. This setup corresponds to two distinct partitions of the dataset. In the first case (\textit{cifar-1}), we sample 500 images per category, resulting in 5,000 training examples to learn the hash mapping. The test set contains 100 images per category (1000 in total). The remaining images are then used to populate the hash table. In the second case (\textit{cifar-2}), we sample 1000 images per category to construct the test set (10,000 in total). The remaining items are both used to learn the hash mapping and populate the hash table. Two images are considered neighbors if they belong to the same class. 

\smallskip
\noindent \textbf{\nus} is a dataset containing 269K images. Each image can be associated with multiple labels, corresponding with 81 ground truth concepts. Following the setup in \cite{Lai_CVPR2015, Zhuang_2016_CVPR, Li_IJCAI2016, Wang_ACCV2016}, we only consider images annotated with the 21 most frequent labels. In total, this corresponds to 195,834 images. The experimental setup also has two distinct partitionings: \textit{nus-1} and \textit{nus-2}. For both cases, a test set is constructed by randomly sampling 100 images per label (2,100 images in total). To learn the hash mapping, 500 images per label are randomly sampled in \textit{nus-1} (10,500 in total). The remaining images are then used to populate the hash table. 
In the second case, \textit{nus-2}, all the images excluding the test set are used in learning the hash mapping and populating the hash table. 
Two images are considered neighbors if they share a single label. We also 
specify a richer neighborhood by allowing multi-level affinities. In this scenario, two images have an affinity value equal to \text{the number of common labels} they share. 

\smallskip
\noindent \textbf{$\mathsf{22K}$ \labelme} consists of 22K images, each represented with a 512-dimensionality $\mathsf{GIST}$ descriptor. Following \cite{BRE, cakir2015AdaptHash}, we randomly partition the dataset into two: a training and test set consisting of 20K and 2K instances, respectively. A 5K subset of the training set is used in learning the hash mapping. As this dataset is unsupervised, we use the $l_2$ norm in determining the neighborhood. Similar to \nus, we allow multi-level affinities for this dataset. We consider four distance percentiles deduced from the training set and assign multi-level affinity values between the instances. 

\smallskip
\noindent \textbf{\imnet} is a subset of ImageNet \cite{deng2009imagenet} containing 130K images from 100 classes. We follow \cite{hashnet} and sample 100 images per class for training. All images in the selected classes from the ILSVRC 2012 validation set are used as the test set. Two images are considered neighbors if they belong to the same class.

\smallskip
Experiments without using multi-level affinities in defining the neighborhood are evaluated using a variant of Mean Average Precision ($\mathsf{mAP}$), depending on the protocol we follow. We collectively group these as \textit{binary affinity} experiments. \textit{Multi-level affinity} experiments are evaluated using Normalized Discounted Cumulative Gain ($\mathsf{NDCG}$), a metric standard in information retrieval for measuring ranking quality with multi-level similarities. In both experiments, Hamming distances are used to retrieve and rank data instances.

We term our method \ecochash~(\textbf{H}ashing with \textbf{B}inary \textbf{M}atrix \textbf{P}ursuit), and compare it against state-of-the-art hashing methods. These methods include: Spectral Hashing (\textbf{SH}) \cite{SH}, Iterative Quantization (\textbf{ITQ}) \cite{ITQ}, 
Supervised Hashing with Kernels (\textbf{SHK}) \cite{SHK},  Fast Hashing with Decision Trees (\textbf{FastHash}) \cite{FastHash}, Structured Hashing (\textbf{StructHash}) \cite{lin2016structured}, Supervised Discrete Hashing (\textbf{SDH}) \cite{Shen_CVPR2015}, Efficient Training of Very Deep Neural Networks (\textbf{VDSH}) \cite{Ziming_VeryDeep_CVPR2016}, Deep Supervised Hashing with Pairwise Labels (\textbf{DPSH}) \cite{Li_IJCAI2016}, Deep Supervised Hashing with Triplet Labels (\textbf{DTSH}) \cite{Wang_ACCV2016} and Mutual Information Hashing (\textbf{MIHash})\cite{mihash, cakir2018hashing}. 
These competing methods have been shown to outperform earlier and other works such as \cite{SSH, MDH, BRE, MLH, Xia_AAAI2014, Lai_CVPR2015, Zhao_CVPR2015}.

For {\cifar} and {\nus} experiments, we fine tune the {\vggf} architecture. For {\imnet} experiments, we fine-tune the {\alex} architecture. Both deep learning models are pretrained using the ImageNet dataset. For \textit{non-deep} methods, we use the output of the penultimate layer of both architectures. For the $\mathsf{22K}$ {\labelme} benchmark, we learn shallow models on top of the {\gist} descriptor. For deep learning based hashing methods, this corresponds to using a single fully connected neural network layer. 


\subsection{Results} 
We provide results for experiments with \textit{binary similarities} with $\mathsf{mAP}$~as the evaluation criterion, and then for \textit{multi-level similarities} with $\mathsf{NDCG}$. In {\cifar}, set $\mathcal{X}$, in which the binary inference is performed upon, represents the 10 classes. For {\nus}, as the neighborhood is defined using the multi-labels, it is then intuitive for set $\mathcal{X}$ to represent label combinations. In our case, we consider unique label combinations in the training set resulting in $\mathcal{X}=4850$ items for binary inference. For the 22K {\labelme} dataset, the items directly correspond to training instances. 
We provide results for the \texttt{regress} binary inference scheme, denoted simply as \ecochash. A comparison between \texttt{constant} and \texttt{regress} is given in the supplementary material.

\begin{table*}[!t]
\footnotesize
\centering
{
\begin{tabular}{l|v|v|v|v|v|v|v|v} 
\hline
$\mathsf{VGG-F}$ & \multicolumn{4}{c|}{ $\mathsf{CIFAR-10}~(\mathsf{mAP})$ } & \multicolumn{4}{c}{ $\mathsf{NUSWIDE}~(\mathsf{mAP@5K})$ } \rule{0pt}{1em} \\
\hline
  \textbf{Method}  & 12 Bits & 24 Bits   & 32 Bits   & 48 Bits & 12 Bits & 24 Bits   & 32 Bits   & 48 Bits \\ 
\hline
SH \cite{SH} & 0.183 & 0.164 & 0.161 & 0.161 & 0.621& 0.616 & 0.615 & 0.612\\
ITQ \cite{ITQ} & 0.237 & 0.246 & 0.255 & 0.261 & 0.719 & 0.739 & 0.747 & 0.756 \\
SHK \cite{SHK} & 0.488 & 0.539 & 0.548 & 0.563 & 0.768 & 0.804& 0.815 & 0.824\\
SDH \cite{Shen_CVPR2015} & 0.478 & 0.557 & 0.584 & 0.592 & \textbf{0.780} & 0.804 & {0.816} & 0.824 \\
FastHash \cite{FastHash} & 0.553 & 0.607 & 0.619 & 0.636 & 0.779 & 0.807 & {0.816} & {0.825}\\
StructHash \cite{lin2016structured} &  {0.664} & {0.693}  & {0.691} & {0.700} &  0.748 & 0.772 & 0.790 & 0.801 \\
VDSH \cite{Ziming_VeryDeep_CVPR2016} & 0.538 & 0.541 & 0.545 & 0.548 & 0.769 & 0.796 & 0.803 & 0.807\\
DPSH \cite{Li_IJCAI2016} & {0.713} & 0.727 & 0.744 & 0.757 & 0.758 & 0.793 & 0.818 & 0.830\\
DTSH \cite{Wang_ACCV2016} & 0.710 & {0.750} & {0.765} & {0.774} & {0.773} & {0.813} & 0.820 & 0.838\\
MIHash \cite{mihash} & 0.738 & 0.775 & 0.791 & 0.816 & 0.773 & \textbf{0.820} & \textbf{0.831} & \textbf{0.843} \\

\ecochash  & \textbf{0.799} & \textbf{0.804} & \textbf{0.830} & \textbf{0.831} & 0.757 & 0.805 & {0.822} & {0.840} \\
\hline
\end{tabular}
}
\caption{Binary affinity experiments on {\cifar} and {\nus} datasets with \textit{cifar-1} and \textit{nus-1} partitionings. The underlying deep learning architecture is \vggf. {\ecochash} outperforms competing methods on {\cifar}, and shows improvements, especially with lengthier codes on {\nus}. 
} 
  \label{table:vggf-setting1}
\end{table*}

\begin{table*}[!t]
\footnotesize
\centering
{
\begin{tabular}{l|v|v|v|v|v|v|v|v} 
\hline
$\mathsf{VGG-F}$ & \multicolumn{4}{c|}{ $\mathsf{CIFAR-10}~(\mathsf{mAP})$ } & \multicolumn{4}{c}{ $\mathsf{NUSWIDE}~(\mathsf{mAP@50K})$ } \rule{0pt}{1em} \\
\hline
    \textbf{Method} & 16 Bits & 24 Bits   & 32 Bits   & 48 Bits & 16 Bits & 24 Bits & 32 Bits   & 48 Bits \\
\hline
DRSH \cite{Zhao_CVPR2015} & 0.608 & 0.611 & 0.617 & 0.618 & 0.609 & 0.618 & 0.621 & 0.631 \\
DRSCH \cite{DSCH} & 0.615 & 0.622 & 0.629 & 0.631 & 0.715 & 0.722 & 0.736 & 0.741 \\
DPSH \cite{Li_IJCAI2016} & 0.903 & 0.885 & 0.915 & 0.911 & 0.715 & 0.722 & 0.736 & 0.741 \\
DTSH \cite{Wang_ACCV2016} & {0.915} & {0.923} & {0.925} & {0.926} & {0.756} & {0.776} & {0.785} & {0.799} \\
MIHash \cite{mihash} & 0.927 & 0.938 & 0.942 & 0.943 & 0.798 & 0.814 & 0.819 & 0.820 \\

\ecochash  & \textbf{0.942} & \textbf{0.944} & \textbf{0.945} & \textbf{0.945} & \textbf{0.804} & \textbf{0.829} & \textbf{0.841} & \textbf{0.855} \\
\hline
\end{tabular}
}
\caption{Binary affinity experiments on {\cifar} and {\nus} datasets with \textit{cifar-2} and \textit{nus-2} partitionings (with {\vggf} architecture). {\ecochash} achieves new state-of-the-art performances, significantly improving over competing methods. 
} 
  \label{table:vggf-setting2}
\end{table*}

\subsubsection{Binary Affinity Experiments.}
Table \ref{table:vggf-setting1} gives results for the \textit{cifar-1} and \textit{nus-1} experimental settings in which $\mathsf{mAP}$ and $\mathsf{mAP@5K}$ values are reported for the {\cifar} and {\nus} datasets, respectively. Deep-learning based hashing methods such as DPSH, DTSH and MIHash outperform most non-deep hashing solutions. This is not surprising as feature representations are simultaneously learned along the hash mapping in these methods. Certain two-stage methods, \eg, FastHash, remain competitive and top deep learning methods including DTSH and MIHash for various hash code lengths, especially for {\nus}. Our two-stage method, {\ecochash}, outperforms all competing methods in majority of the cases, including MIHash, DTSH and DPSH with very large improvement margins. Specifically for {\cifar}, the best competing method is MIHash, a recent study that learns the hash mapping using a mutual information formulation. The improvement over MIHash is over $\mathbf{6\%}$ for certain hash code lengths, \eg, for 12 bits $\mathbf{0.799}$ \vs $0.738$ $\mathsf{mAP}$. Our method significantly improves over SHK as well, which also proposes a matrix fitting formulation but learns its hash mapping in an interleaved manner. This validates defining the binary code inference over items that directly define the neighborhood, {\ie} classes for {\cifar}.

For the {\nus} dataset, the binary inference is done over the set of label combinations in the training data. {\ecochash} demonstrates either comparable results or outperforms the state-of-the-art hashing methods. 
A relevant recent two-stage hashing method is \cite{Zhuang_2016_CVPR} in which the same settings (\textit{cifar-1} and \textit{nus-1}) are used but with fine-tuning a $\mathsf{VGG-16}$ architecture. Their {\cifar} and {\nus} results have at most $0.80$ $\mathsf{mAP}$ and $0.75$ $\mathsf{mAP@5K}$ values, respectively, for all hash code lengths. {\ecochash}, on the other hand, achieves these performance values  with the inferior {\vggf} architecture. 

To further emphasize the merits of {\ecochash}, we consider the experimental settings \textit{cifar-2} and \textit{nus-2} and compare against recent deep-learning hashing methods. In this setting, we again fine-tune the {\vggf} architecture pretrained on ImageNet. Table \ref{table:vggf-setting2} gives the results. Notice that our method significantly outperforms all techniques, and yields new state-of-the-art results for {\cifar} and {\nus}. 

Retrieval results for {\imnet} are given in Table \ref{table:alexnet-imagenet}. In these experiments, we only compare against MIHash, the overall best competing method in past experiments and HashNet \cite{hashnet}, another very recent deep learning based hashing study. As demonstrated, {\ecochash} establishes the new state-of-the-art in image retrieval for this benchmark. {\ecochash} outperforms both methods significantly, {\eg}, with 64-bits, we demonstrate $4-6\%$ improvement. This further validates the quality of the binary codes produced with {\ecochash}.


\begin{table}[!t]
\footnotesize
\centering
{
\begin{tabular}{l|v|v|v|v} 
\hline
$\mathsf{AlexNet}$ & \multicolumn{4}{c}{ $\mathsf{ImageNet100}~(\mathsf{mAP@1K})$ } \rule{0pt}{1em} \\
\hline
  \textbf{Method}  & 16 Bits & 32 Bits   & 48 Bits   & 64 Bits  \\
\hline
HashNet \cite{hashnet}  & 0.506 &  0.630 & 0.663 & 0.683 \\
MIHash \cite{mihash}  & 0.569 &  0.661 & 0.685 & 0.694 \\

\ecochash   & \textbf{0.574} & \textbf{0.692} & \textbf{0.712} & \textbf{0.742} \\
\hline
\end{tabular}
}
\caption{$\mathsf{mAP@1K}$ values on {\imnet} using {\alex}. {\ecochash} outperforms the two state-of-the-art formulations using mutual information \cite{mihash} and continuation methods \cite{hashnet}. 
} 
  \label{table:alexnet-imagenet}
\end{table}

\subsubsection{Multilevel Affinity Experiments.}
In these experiments, we allow multi-level similarities between items of set $\mathcal{X}$ and use $\mathsf{NDCG}$ as the evaluation criterion. For {\nus}, we consider the number of shared labels as affinity values. For $\mathsf{22K}$ {\labelme} dataset, we consider using distance percentiles $\{2\%, 5\%, 10\%, 20\%\}$ deduced from the training set to assign inversely proportional affinity values between the training instances. This emphasizes multi-level rankings among neighbors in the original feature space. In $\mathsf{22K}$ {\labelme}, we use a single fully connected layer as the hash mapping for the deep-learning based methods.  

Table \ref{table:multilevel} gives results. For {\nus}, {\ecochash} outperforms all state-of-the-art methods 
including MIHash. 
In $\mathsf{22K}$ {\labelme}, {\ecochash} either achieves state-of-the-art performance, or is a close second. 
An interesting observation is that, when the feature learning aspect is removed due to the use of precomputed \gist~features, non-deep methods such as FastHash and StructHash outperform deep-learning hashing methods DPSH and DTSH. 
While FastHash and StruchHash enjoy non-linear hash functions such as boosted decision trees, this also indicates that the prowess of DPSH and DTSH might come primarily through feature learning. 
On the other hand, both {\ecochash} and MIHash show top performances with a single fully connected layer as the hash mapping, indicating that they produce binary codes that more accurately reflect the neighborhood. Regarding $\mathsf{22K}$ {\labelme}, for {\ecochash}, set $\mathcal{X}$ corresponds to training instances, as similarly in other methods. This suggests that the performance improvement of {\ecochash} is not merely due to the fact that the binary inference is performed upon items that directly define the neighborhood, but also due to our formulation that learns a Hamming metric with optimally selected bit weights.

\begin{table*}[t]
\footnotesize
\centering
{
\begin{tabular}{l|v|v|v|v|v|v|v|v} 
\hline
 & \multicolumn{4}{c|}{ $\mathsf{NUSWIDE}~(\mathsf{VGG-F}, \mathsf{NDCG})$ } & \multicolumn{4}{c}{ $\mathsf{22K~LabelMe}~(\mathsf{GIST}, \mathsf{NDCG})$ } \rule{0pt}{1em} \\
\hline
 \textbf{Method}   & 16 Bits & 32 Bits   & 48 Bits   & 64 Bits & 16 Bits & 24 Bits & 32 Bits   & 48 Bits \\
\hline
FastHash \cite{FastHash}  & {0.885} & {0.896}  & {0.899} & {0.902} 
& 0.672 & 0.716 & 0.740 & 0.757 \\ 
StructHash \cite{lin2016structured} & 0.889 & 0.893 & 0.894 & 0.898
& 0.704 & 0.768 & 0.802 & 0.824 \\ 
DPSH \cite{Li_IJCAI2016}  & 0.895 & 0.905 & 0.909 & 0.909 
& 0.677 & 0.740 & 0.755 & 0.765  \\ 
DTSH \cite{Wang_ACCV2016}  & 0.896 &  0.905 & 0.911  & 0.913  
& 0.620 & 0.685 & 0.694 & 0.702 \\ 
MIHash \cite{mihash}  & 0.886 &  0.903 & 0.909  & 0.912 
&  0.713 & 0.822 & \textbf{0.855} & \textbf{0.873} \\ 

\ecochash  & \textbf{0.914} & \textbf{0.924} & \textbf{0.927} & \textbf{0.930}  & \textbf{0.823} & \textbf{0.829} & 0.849 & 0.866 \\
\hline
\end{tabular}
}
\caption{Multi-level affinity experiments on {\nus} and $\mathsf{22K}$ {\labelme} using {\vggf} and {\gist}, respectively. The partitioning used for {\nus} is \textit{nus-1}. The evaluation criterion is Normalized Discounted Cumulative Gain ($\mathsf{NDCG})$. {\ecochash} improves over the state-of-the-art in majority of the cases. 
} 
  \label{table:multilevel}
\end{table*}

\section{Conclusion}
We have proposed improvements to a commonly used formulation in two-stage hashing methods. 
We first provided a theoretical result on the quality of the binary codes showing that, under mild assumptions, we can construct binary codes that fit the neighborhood with arbitrary accuracy. Secondly, we analyzed the sub-optimality of binary codes constructed as to fit an affinity matrix that is not defined on items directly related to the neighborhood. Incorporating our findings, we proposed a novel two-stage hashing method that significantly outperforms previous hashing studies on multiple benchmarks.

\subsubsection*{Acknowledgments.}
The authors thank Sarah Adel Bargal for helpful discussions. 
This work is primarily conducted at Boston University, supported in part by a BU IGNITION award, and equipment donated by NVIDIA.

\bibliographystyle{unsrt}
\bibliography{egbib}

\section*{A. Proofs}
\label{appendix-a}
We provide the arguments for convenience.

\medskip
\noindent \textbf{Property 1.} \textit{Let $Q_{t}$ be the residual $-\nabla f(\mathcal{U}_{t})$ at iteration $t$. There exists a $\mathcal{R}$ such that $\forall t, \|Q_{t}\|_F > 0$}.%

\medskip
\noindent \textit{Proof.} Given the fact that $\mathbb{Z}$ is closed under addition, and since $\mathbf{v}\mathbf{v}^\top \in \mathbb{Z}$, we have $\sum_t \mathbf{v}_t \mathbf{v}_t^\top \in \mathbb{Z}$. If matrix $\mathcal{R}$ has an element such that $\mathcal{R}_{ij} \in \mathbb{R} \backslash \mathbb{Z}$ then $\forall t, \|Q_{t}\|_F > 0$.
\hfill $\square$

\medskip
To prove \text{Theorem 2}, we provide Property 3. For this purpose, assume $\mathbf{v}_t = \text{sgn}(\mathbf{v}^*)$ is the approximate solution in Eq.~\ref{gradient} used in our gradient descent update rule, where $\mathbf{v}^*$ is the solution of the Rayleigh Quotient in Eq.~\ref{rayleigh}. The following property bounds the value of $\langle \mathbf{v}_t \mathbf{v}_t^\top, Q_{t-1}\rangle$.

\medskip
\noindent \textbf{Property 3.} \textit{If  $\mathbf{v}_t = \text{sgn}(\mathbf{v}^*)$ is the approximate solution in Eq.~\ref{gradient} where $\mathbf{v}^*$ is the solution of the Rayleigh Quotient in Eq.~\ref{rayleigh}, then $n\lambda_{\min}(Q_{t-1}) \leq \langle \mathbf{v}_t \mathbf{v}_t^\top, Q_{t-1}\rangle \leq n\lambda_{\max}(Q_{t-1})$ where $\lambda_{\min}(Q_{t-1})$ and $\lambda_{\min}(Q_{t-1})$ denote the smallest and largest eigenvalues of $Q_{t-1}$, respectively}. 

\medskip
\noindent \textit{Proof.} It can be shown that the value of Rayleigh Quotient $R(Q) = \frac{\mathbf{v}^\top Q\mathbf{v}}{\mathbf{v}^\top\mathbf{v}} $ has range $[\lambda_{\min}(Q), \lambda_{\max}(Q)]$ where $Q$ is real symmetric matrix  and $\mathbf{v}$ is any non-zero vector. Since:
\begin{equation}
\begin{split}
n \lambda_{\min}(Q) = \min_{\mathbf{v}^\top \mathbf{v} = n} \mathbf{v}^\top Q\mathbf{v} \leq & \min_{\mathbf{v} \in \{-1,+1\}^n} \mathbf{v}^\top Q\mathbf{v} \leq \\
& \max_{\mathbf{v} \in \{-1,+1\}^n} \mathbf{v}^\top Q\mathbf{v} \leq \\
& \max_{\mathbf{v}^\top \mathbf{v} = n} \mathbf{v}^\top Q\mathbf{v} = n\lambda_{\max}(Q), 
\end{split} 
\end{equation}
the value of $\mathbf{v}_t^\top Q \mathbf{v}_t$ where $\mathbf{v}_t = \text{sqn}(\mathbf{v}^*)$ is within the range $[n \lambda_{\min}(Q), n \lambda_{\max}(Q) ]$. 

\hfill $\square$

\medskip
\noindent Now we're ready to prove Theorem 2. 

\medskip
\noindent \textbf{Theorem 2.} \textit{If $\alpha_t \in \mathbb{R}$, then the gradient descent algorithm Eq.~\ref{sgd}-\ref{gradient} satisfies}
\begin{equation}
\|Q_t\|_F \leq \eta^{t-1}\|Q_{t-1}\|_F,~~\forall t
\label{theorem-supp}
\end{equation}
\textit{where $\eta \in [0, 1]$}. 

\medskip
\noindent \textit{Proof.} If $d_{\max}$ is bounded then $\|\nabla f(\mathcal{U})\|_F \leq L$ for some constant $L$. Eq.~\ref{main_obj3} is then a $L$-Lipschitz continuous function. From the Lipschitz definition: 
\begin{equation}
f(\mathcal{U}_t) \leq f(\mathcal{U}_{t-1}) + \langle \mathcal{U}_t - \mathcal{U}_{t-1}, \nabla f(\mathcal{U}_{t-1})\rangle + \frac{L}{2} \|\mathcal{U}_t - \mathcal{U}_{t-1}\|_F^2 
\end{equation}

\noindent Let $r = \langle \mathbf{v}_t\mathbf{v}_t^\top, \nabla f(\mathcal{U}_{t-1}) \rangle$. Since $\mathcal{U}_{t} - \mathcal{U}_{t-1} = \alpha_t \mathbf{v}_t \mathbf{v}_t^\top$ and $\|\mathbf{v}_t \mathbf{v}_t^\top\|_F^2 = n^2$ we have:
\begin{equation}
f(\mathcal{U}_t) \leq f(\mathcal{U}_{t-1}) + \alpha_t \langle \mathbf{v}_t\mathbf{v}_t^\top, \nabla f(\mathcal{U}_{t-1})\rangle + \frac{\alpha_t^2 n^2 L }{2} 
\end{equation}

\noindent Select $\alpha_t \in [0, -\frac{2r}{n^2 L }]$. Without loss of generality set $\alpha_t = -\frac{r}{n^2L}$. We then have:
\begin{equation}
f(\mathcal{U}_t) \leq f(\mathcal{U}_{t-1}) - \frac{r^2}{2n^2L }.
\label{before_final}
\end{equation}
\noindent Let 
\begin{equation}
r = \langle \mathbf{v}_t\mathbf{v}_t^\top, \nabla f(\mathcal{U}_{t-1}) \rangle = \beta_{t-1} \|\nabla f(\mathcal{U}_{t-1})\|_F \|\mathbf{v}_t \mathbf{v}_t^\top\|_F  
\end{equation}
where $\beta_{t-1} = \frac{\langle \mathbf{v}_t\mathbf{v}_t^\top, \nabla f(\mathcal{U}_{t-1}) \rangle}{\|\nabla f(\mathcal{U}_{t-1})\|_F \|\mathbf{v}_t \mathbf{v}_t^\top\|_F}$. 
Substituting $r$ in the above equation, we have:
\begin{equation}
f(\mathcal{U}_t) \leq f(\mathcal{U}_{t-1}) - \frac{\beta_{t-1}^2}{2L} \| \nabla f(\mathcal{U}_{t-1})\|_F^2
\end{equation}
Since $f(\mathcal{U}) = \frac{1}{2} \|\nabla f(\mathcal{U})\|_F^2$ we have: 
\begin{equation}
\|\nabla f(\mathcal{U}_t)\|_F^2 \leq (1 - \frac{\beta_{t-1}^2}{L}) \|\nabla f(\mathcal{U}_{t-1})\|_F^2 
\end{equation}
Using Property 3 we can bound $\beta_{t-1}^2$ as:
\begin{equation}
\begin{split}
\frac{n\lambda_{\min}(\nabla f(\mathcal{U}_{t-1}))}{\|\nabla f(\mathcal{U}_{t-1})\|_F \|\mathbf{v}_t \mathbf{v}_t^\top\|_F} & \leq \beta_{t-1} \leq \frac{n\lambda_{\max}(\nabla f(\mathcal{U}_{t-1}))}{\|\nabla f(\mathcal{U}_{t-1})\|_F \|\mathbf{v}_t \mathbf{v}_t^\top\|_F}  
\end{split} 
\end{equation}
which yields:
\begin{equation}
\begin{split}
0 & \leq \beta_{t-1}^2 \leq \frac{\max (\lambda_{\max}^2(\nabla f(\mathcal{U}_{t-1})) , \lambda_{\min}^2(\nabla f(\mathcal{U}_{t-1})))}{\|\nabla f(\mathcal{U}_{t-1})\|_F^2}  \\
0 & \leq \beta_{t-1}^2 \leq 1 
\end{split}
\end{equation}
where we use the facts that $|\lambda(A)| \leq \|A\|_F, ~\forall \lambda(A)$ where $\lambda (A)$ denotes the eigenvalue of matrix $A$, and $\|\mathbf{v}_t \mathbf{v}_t^\top\|_F = n$, 
Since $L > 1$ by construction and $Q_t = -\nabla f(\mathcal{U}_{t})= \mathcal{R} - \sum_t \alpha_t \mathbf{v}_t\mathbf{v}_t^\top$ equals the residual, this concludes our proof. 

\hfill $\square$


\medskip
\noindent \textbf{Corollary 3.} \textit{If $\mathbf{v}_t^\top Q_{t-1} \mathbf{v}_t \neq 0, ~\forall t$ then the residual norm $\|Q_t\|_F$ strictly decreases}.

\medskip
\noindent \textit{Proof.} Notice that $\mathbf{v}_t^\top Q_{t-1}\mathbf{v}_t \neq 0$ implies $\beta_{t-1}^2 \in (0, 1]$. Then $\eta \in [0,1)$ in Eq.~\ref{theorem}.

\hfill $\square$

\medskip
We now provide Property 4 stating that {refinement} of the step-sizes does not break the monotonicity as defined in Theorem 2 and Corollary 3. 

\medskip
\noindent \textbf{Property 4.} \textit{Let $Q_t$ be the residual matrix at iteration $t$ and $\alpha_t$ set according to Theorem 2. Let $\widehat{Q}_t$ be the residual after refining the step-sizes $\bm{\alpha_t} = [\alpha_1, \cdots, \alpha_t]^\top$ using Eq.~\ref{solution-regression}. Then $\|\widehat{Q}_t\|_F \leq \|Q_t\|_F$}.

\medskip
\noindent \textit{Proof.} Eq.~\ref{regression} has a closed form solution and its minimum is achieved when $\bm{\alpha}_t$ is set according to Eq.~\ref{solution-regression}. Since the search space of Eq.~\ref{regression} contains the initial step-size values, the inequality $\|\widehat{Q}_t\|_F \leq \|Q_t\|_F$ holds. 
\hfill $\square$

\section*{B. Constant \vs Regress Experiments}
Table \ref{table:constant-regress-binary-affinity} - \ref{table:constant-regress-nuswide-ndcg} provides retrieval results for the \texttt{constant} {\vs} \texttt{regress} binary inference scheme experiments. One observation is that, for multi-class datasets {\cifar} and {\imnet}, \texttt{constant} and \texttt{regress} show similar retrieval performances. This is not surprising as the standard definition of the neighborhood for these two datasets consider two instances to be neighbors if they belong to the same class \cite{Lai_CVPR2015,Zhuang_2016_CVPR,Li_IJCAI2016,Wang_ACCV2016}. With this neighborhood, distances between pairs of classes are all equal, consequently the step sizes merely become a scaling factor for the hamming distances between the pairs.    

On the other hand, when the neighborhood definition involves a \textit{richer} neighborhood, weighted Hamming distances (\texttt{regress}) performs better then the ordinary hamming distance, as confirmed with the experiments on {\nus} and $\mathsf{22K}$ {\labelme}. In both binary and multilevel {\nus} experiments, \texttt{regress} achieves significant improvements over \texttt{constant}. This is also true with lengthier codes for $\mathsf{22K}$ \labelme, as shown in Table \ref{table:constant-regress-nuswide-ndcg}.
\begin{table}[h]
\footnotesize
\centering
{
\begin{tabular}{l|i|i|i|i} 
\hline
$\mathsf{VGG-F}$ & \multicolumn{4}{c}{ $\mathsf{CIFAR-10}~(\mathsf{mAP})$ }  \\
\hline
  \textit{cifar-1}  & 16 Bits & 32 Bits   & 48 Bits   & 64 Bits  \\
\hline
\texttt{constant}  & 0.766 & \textbf{0.818} & 0.824 & \textbf{0.839} \\
\texttt{regress}   & \textbf{0.799} & 0.804 & \textbf{0.830} & 0.831 \\
\hline
$\mathsf{VGG-F}$ & \multicolumn{4}{c}{ $\mathsf{NUSWIDE}$~($\mathsf{mAP@5K}$)}\\
\hline
  \textit{nus-1}  & 16 Bits & 32 Bits   & 48 Bits   & 64 Bits  \\
\hline
\texttt{constant}  & 0.675 & 0.723 & 0.731 & 0.773 \\
\texttt{regress}   & \textbf{0.757} & \textbf{0.805} & \textbf{0.822} & \textbf{0.840} \\
\hline
$\mathsf{VGG-F}$ & \multicolumn{4}{c}{ $\mathsf{CIFAR-10}~(\mathsf{mAP})$ }\\
\hline
  \textit{cifar-2}  & 16 Bits & 32 Bits   & 48 Bits   & 64 Bits  \\
\hline
\texttt{constant}  & 0.941 & \textbf{0.944} & \textbf{0.945} & \textbf{0.946} \\
\texttt{regress}   & \textbf{0.942} & \textbf{0.944} & \textbf{0.945} & 0.945 \\
\hline
$\mathsf{VGG-F}$ & \multicolumn{4}{c}{ $\mathsf{NUSWIDE}~(\mathsf{mAP@50K})$ }\\
\hline
  \textit{nus-2}  & 16 Bits & 32 Bits   & 48 Bits   & 64 Bits  \\
\hline
\texttt{constant}  & 0.746 & 0.754 & 0.754 & 0.779 \\
\texttt{regress}   & \textbf{0.804} & \textbf{0.829} & \textbf{0.841} & \textbf{0.855} \\
\hline
$\mathsf{AlexNet}$ & \multicolumn{4}{c}{ $\mathsf{ImageNet100}~(\mathsf{mAP@1K})$ }  \\
\hline
    & 16 Bits & 32 Bits   & 48 Bits   & 64 Bits  \\
\hline
\texttt{constant}  & 0.552 & 0.685 & 0.703 & \textbf{0.751} \\
\texttt{regress}   & \textbf{0.574} & \textbf{0.692} & \textbf{0.712} & 0.742 \\
\hline
\end{tabular}
}
\caption{\texttt{constant} \vs \texttt{regress} results with binary affinities.
} 
  \label{table:constant-regress-binary-affinity}
\end{table}

\begin{table}[t]
\footnotesize
\centering
{
\begin{tabular}{l|i|i|i|i}
\hline
$\mathsf{VGG-F}$ & \multicolumn{4}{c}{ $\mathsf{NUSWIDE}~(\mathsf{NDCG})$ } \rule{0pt}{1em} \\
\hline
    & 16 Bits & 32 Bits   & 48 Bits   & 64 Bits  \\
\hline
\texttt{constant} & 0.905 & 0.910 & 0.909 & 0.916 \\

\texttt{regress} & \textbf{0.914} & \textbf{0.924} & \textbf{0.927} & \textbf{0.930} \\
\hline
\end{tabular}
}
\caption{\texttt{constant} \vs \texttt{regress} results with multi-level affinities on {\nus}.
}
{
\begin{tabular}{l|i|i|i|i|i|i}
\hline
$\mathsf{GIST}$ & \multicolumn{6}{c}{ $\mathsf{22K~LabelMe}~(\mathsf{NDCG})$ } \rule{0pt}{1em} \\
\hline
    & 16 Bits & 24 Bits & 32 Bits & 48 Bits  & 64 Bits & 96 Bits \\
\hline
\texttt{constant} & \textbf{0.837} & \textbf{0.845} & \textbf{0.852} & 0.857 & 0.870 & 0.865 \\

\texttt{regress} & 0.823 & 0.829 & 0.849 & \textbf{0.866} & \textbf{0.892} & \textbf{0.917} \\
\hline
\end{tabular}
}
\caption{\texttt{constant} \vs \texttt{regress} results with multi-level affinities on $\mathsf{22K}$ {\labelme}.
} 

\label{table:constant-regress-nuswide-ndcg}
\end{table}


\section*{C. Constant \vs Regressed Binary Code Inference}
\begin{figure}[t]
\centering
\subfloat{{\includegraphics[clip, trim=2.8cm 6cm 3.0cm 6cm, width=0.24\textwidth]{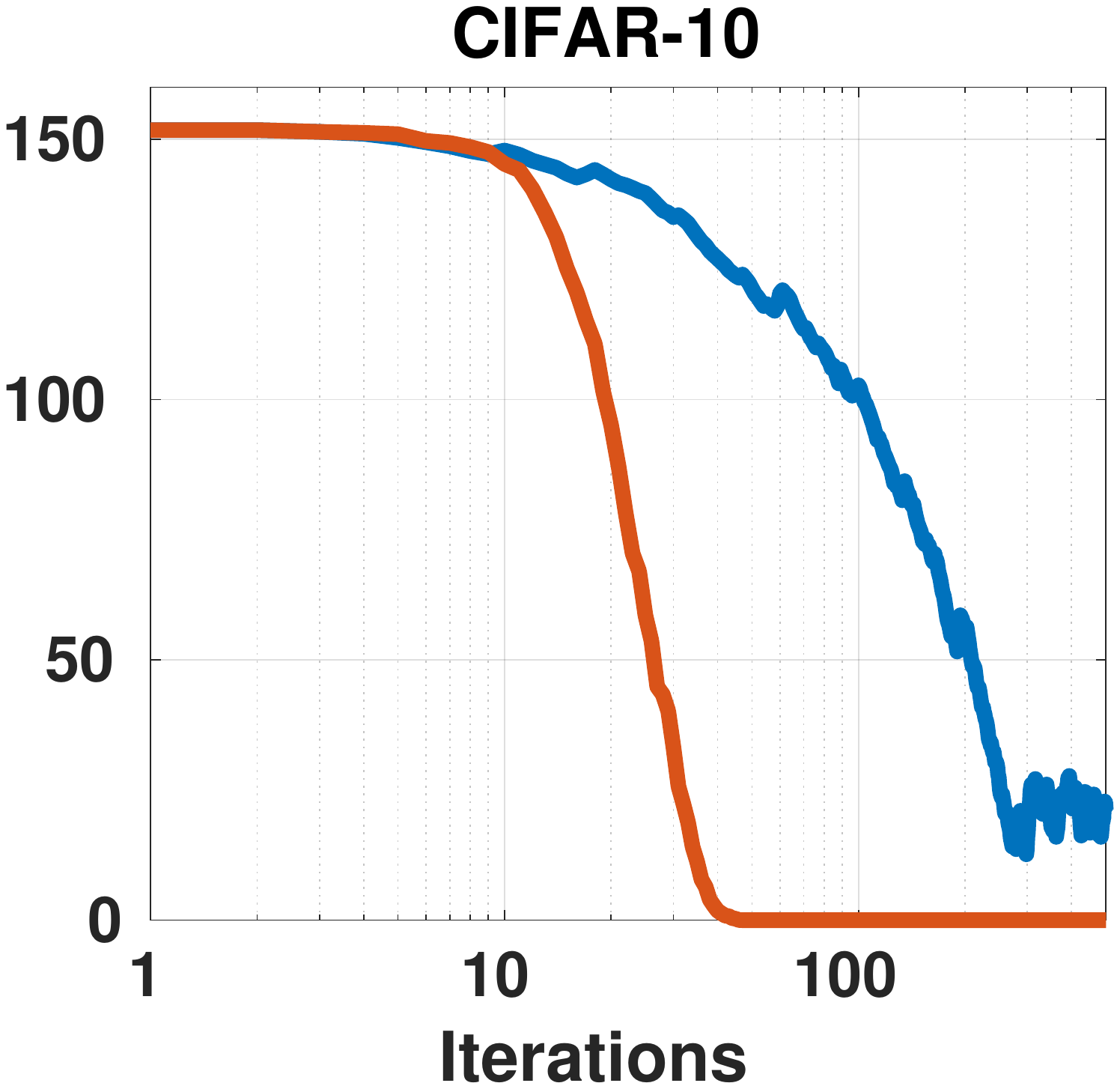}}}
\subfloat{{\includegraphics[clip, trim=2.8cm 6cm 3.0cm 6cm, width=0.24\textwidth]{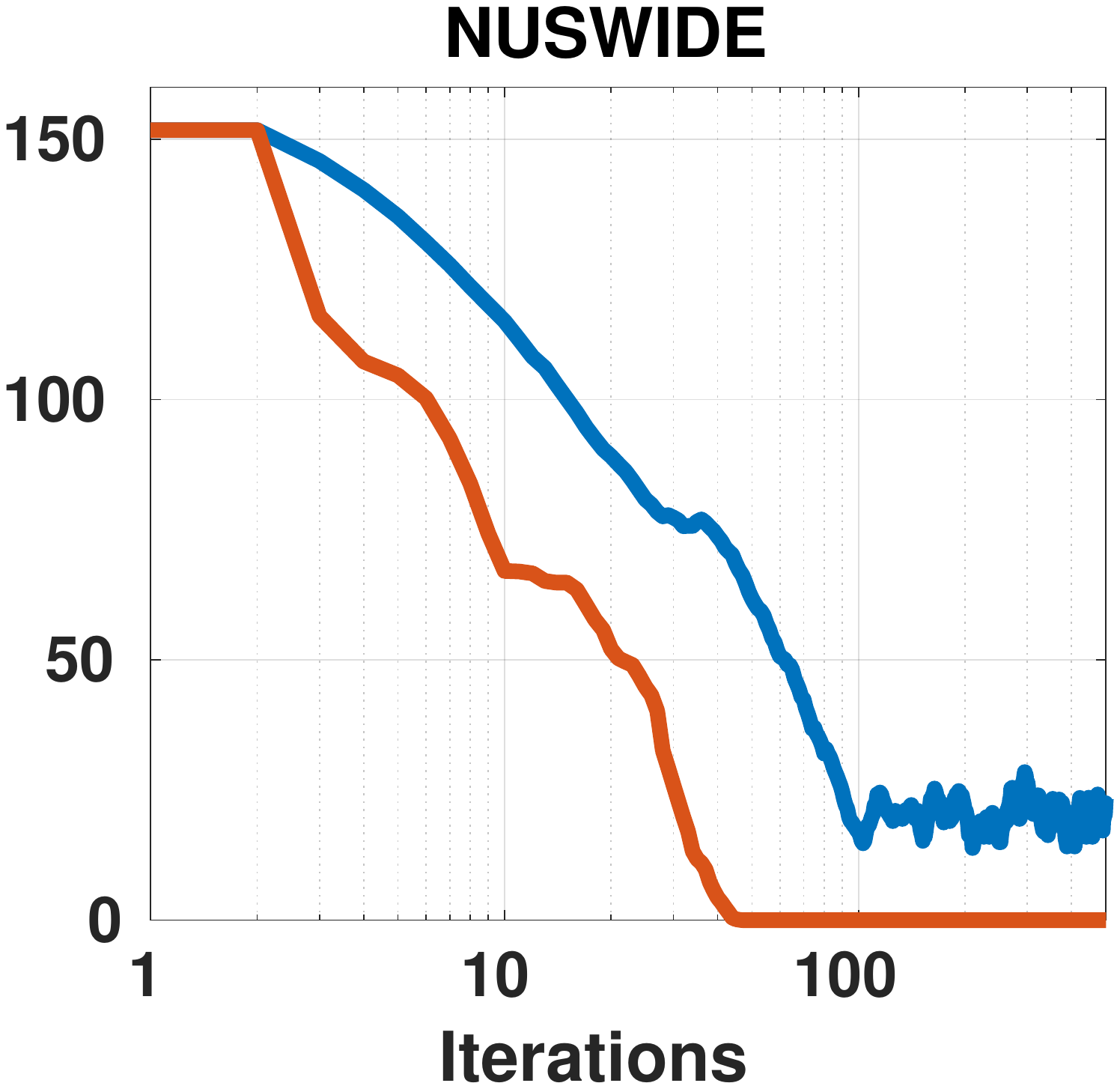}}}
\subfloat{{\includegraphics[clip, trim=2.8cm 6cm 3.0cm 6cm, width=0.24\textwidth]{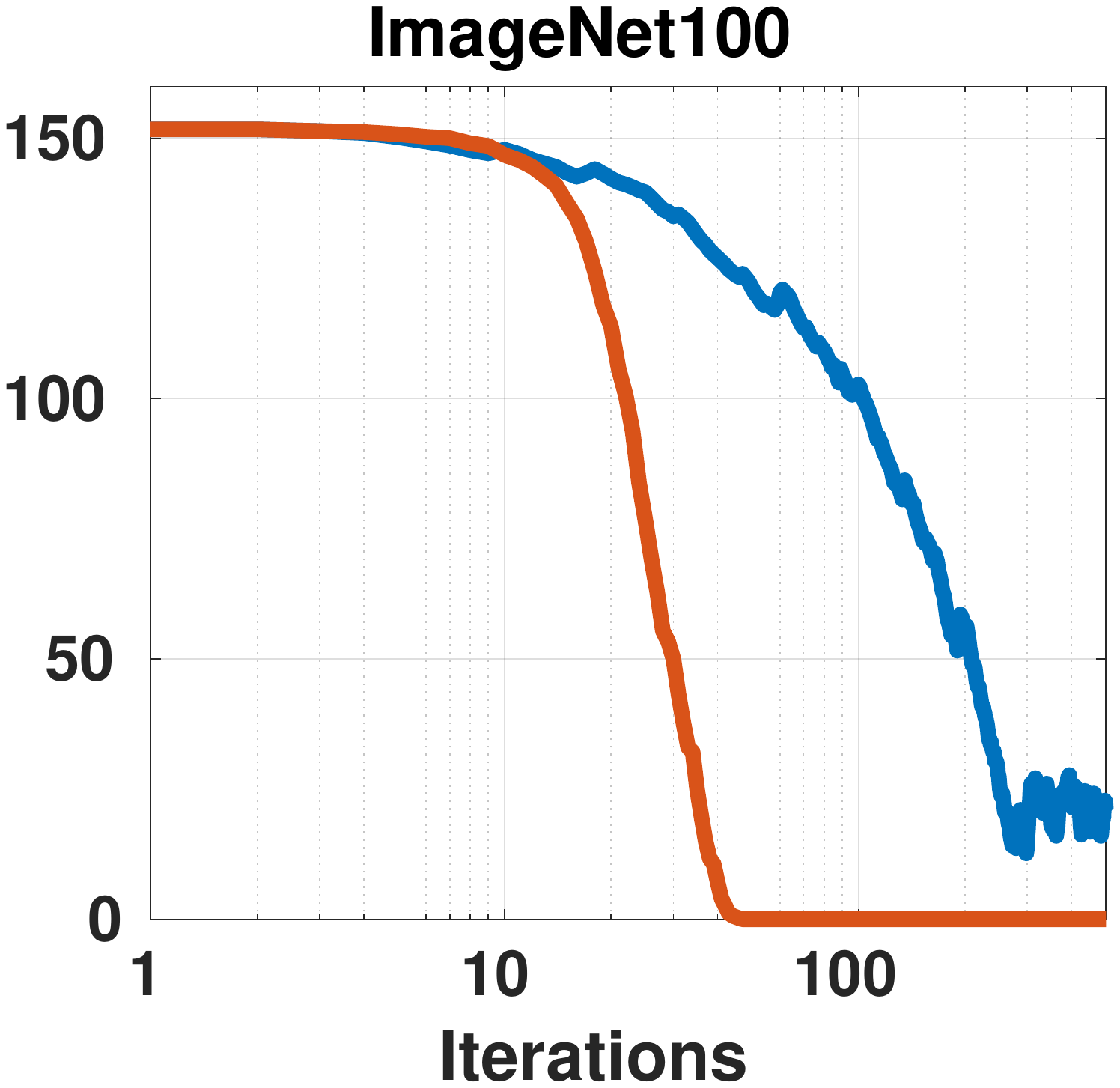}}}
\subfloat{{\includegraphics[clip, trim=2.8cm 6cm 3.0cm 6cm, width=0.24\textwidth]{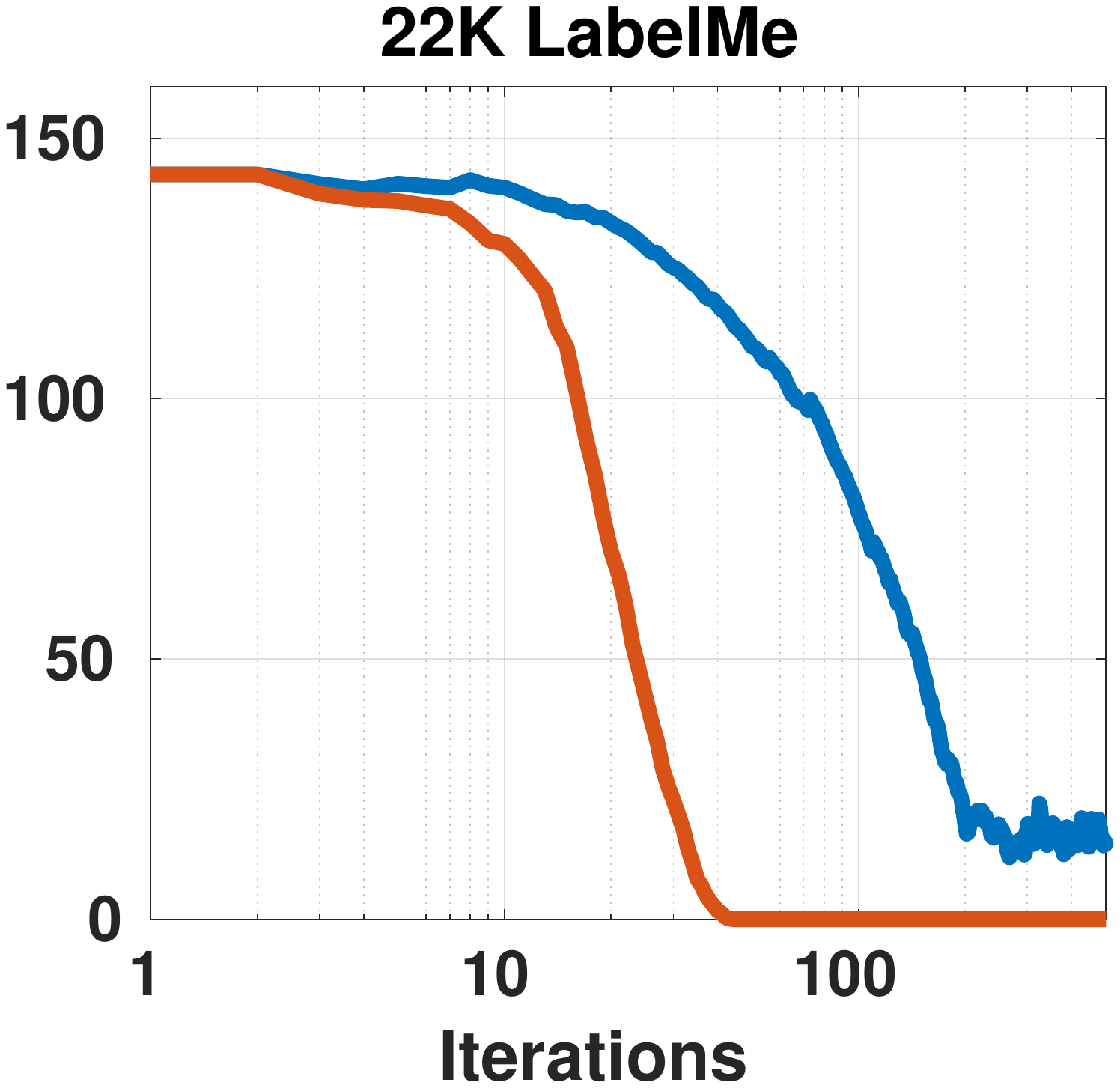}}}
\caption{Norm of the residual matrix \vs iteration, when step sizes in Alg. Binary code inference are \texttt{constant} (\blueline) and \texttt{regress} (\redline). Notice the residual norm of the \texttt{regress} inference scheme converges both faster and to zero.} 
\label{fig:constant-vs-regressed}
\end{figure}

Fig. \ref{fig:constant-vs-regressed} plots the Frobenius norm of the residual matrix $Q_t$ as a function of $t$ for all datasets, and for the \texttt{constant} and \texttt{regress} binary code inference schemes. The neighborhood reflects the binary affinity experiments for all datasets excluding $\mathsf{22K}$ \labelme, in which the default multi-level setup is considered. For convenience, we sample ten items from $\mathcal{X}$ for the inference. Note that as $\mathbf{v}\mathbf{v}^\top$ has all diagonal entries equal to $1$, for $\texttt{constant}$ this implies that the norm will be dominated by the diagonal elements of the residual matrix at further iterations; hence in this figure, we do not consider diagonal terms in computing the norm.

We make two critical observations: the residual norm converges much more faster, and to \textit{zero} for the regressed scheme, justifying our main theorem. While the binary codes constructed with \texttt{constant} fits the neighborhood to some degree, weighted binary codes (\texttt{regress}) allow perfect fit for all datasets.

\section*{D. Visualization of Binary Codes} 
\begin{figure}
\centering
\subfloat{{\includegraphics[clip, width=0.4\textwidth]{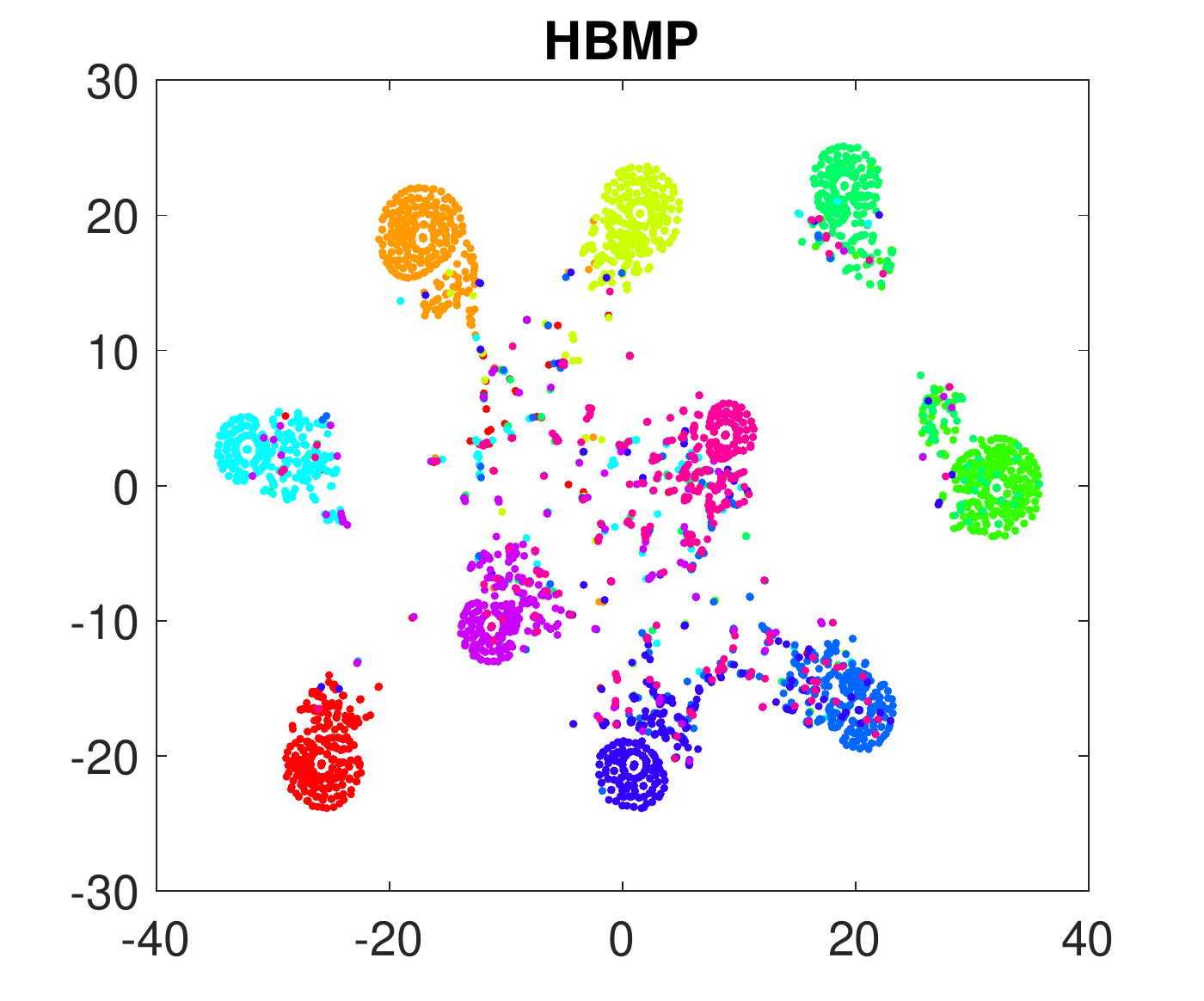}}}
\qquad
\subfloat{{\includegraphics[clip, width=0.4\textwidth]{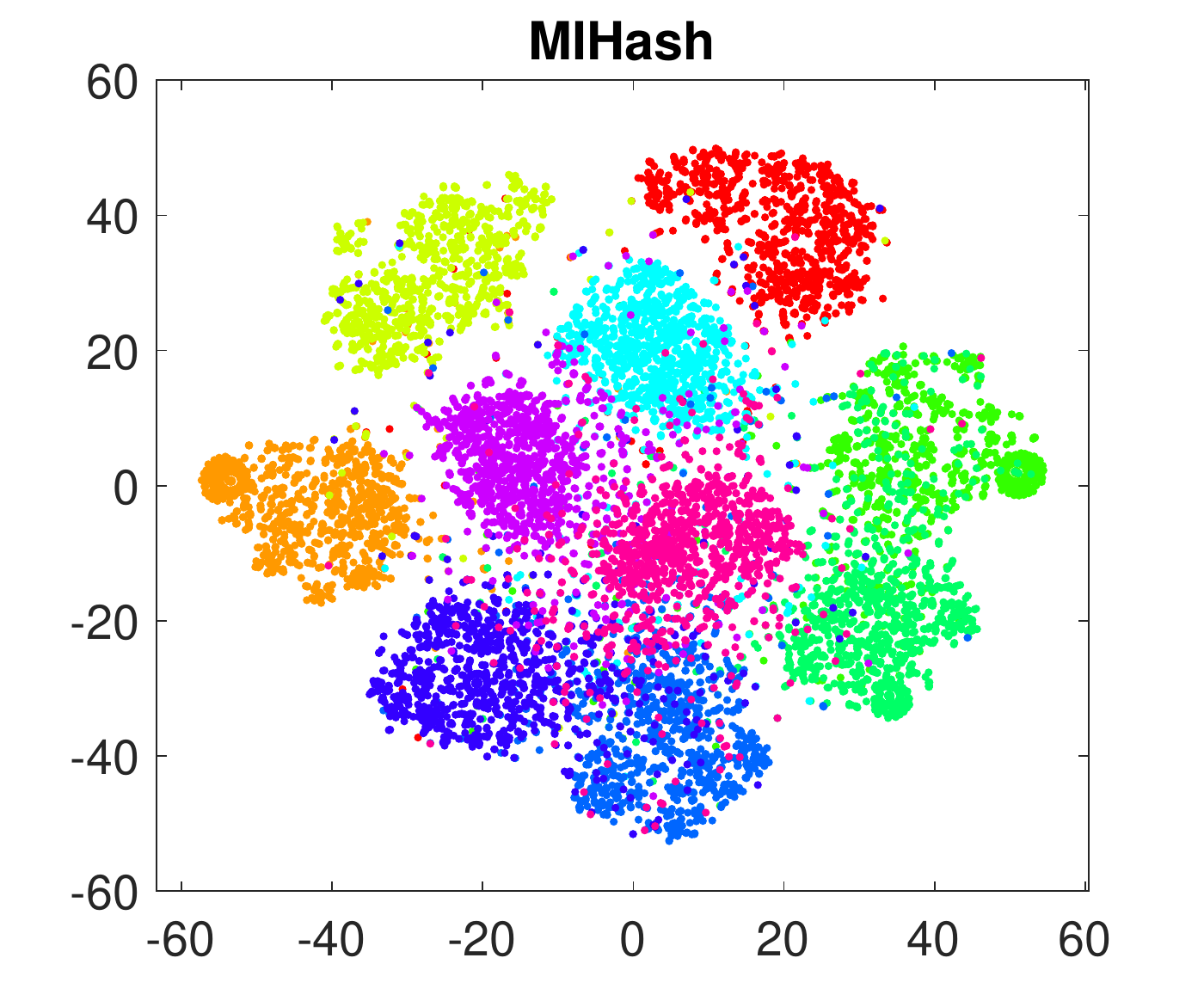}}}
\caption{t-SNE \cite{tsne} visualization of the $48$-bit binary codes produced by {\ecochash} and MIHash on {\imnet}. {\ecochash} yields well separated codes with distinct structures, opposed to MIHash, in which the binary codes have a higher overlap.} 
\label{fig:tsne}
\end{figure}

Fig. \ref{fig:tsne} plots the t-SNE \cite{tsne} visualization for the binary codes constructed by {\ecochash} and the top competing method, MIHash, on {\imnet} (for ease in visualization, we sample 10 categories). Notice that {\ecochash} produces binary codes, well separated into distinct classes. On the other hand, binary codes generated with MIHash have less distinct structures. This correlates well with the formulation of {\ecochash}, where target codes are first generated as to preserve the neighborhood. For {\imnet}, this corresponds to target codes having high separation in the Hamming space. On the other hand, MIHash does not specifically optimize for such a criterion but merely tries to reduce overlaps between the classes. 

\section*{E. Example Retrieval Results} 
\begin{figure}[t]
  \centering
   \includegraphics[width=0.85\textwidth]{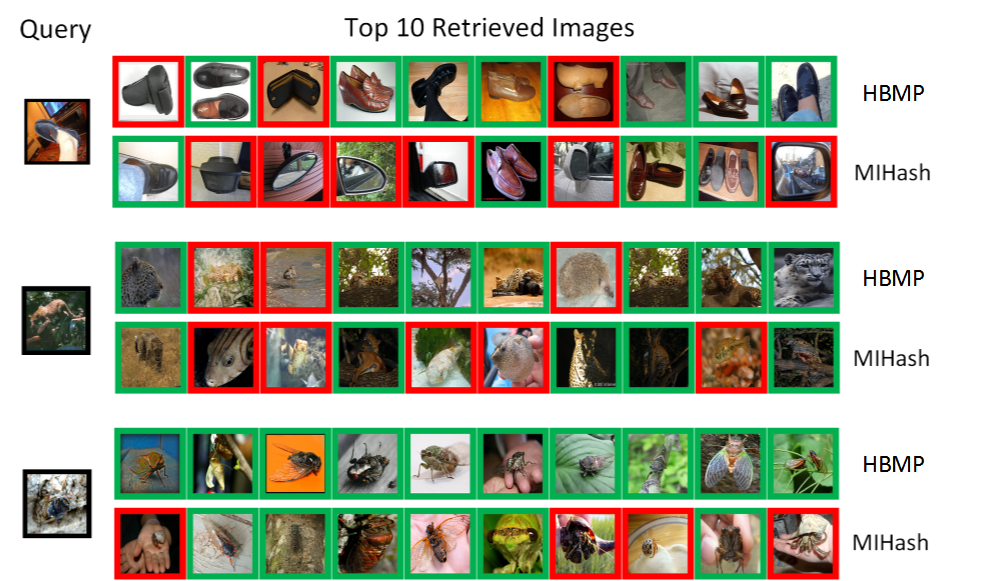}
\caption{Example retrieval results for {\ecochash} and MIHash on {\imnet}. In all the queries, {\ecochash} retrieves more relevant images compared to MIHash. Neighbor and non-neighbor images are denoted with a green and red boundary, respectively.}
\label{fig:retrieval}
\end{figure}

In Fig. \ref{fig:retrieval}, we present example retrieval results for {\ecochash} and MIHash for several image queries from the {\imnet} dataset. The top 10 retrievals of three query images from three distinct categories are presented. Non-neighbors are in red. Most of the retrieved images belong to the same class with the query image. As shown, many of the retrieved images that do not belong to the same class appear visually similar. However, {\ecochash}, retrieves fewer non-neighbor images compared to MIHash. For example, MIHash returns side-mirror images for the first query, and different species images for the third query. Overall, {\ecochash} retrieves much more relevant images. 


%
%
%

%




\end{document}